\documentclass{article}
\usepackage[accepted]{icml2016}
\usepackage{times}
\usepackage{natbib}
\usepackage{graphicx}
\usepackage{color}
\usepackage{tikz}
\usetikzlibrary{arrows,shapes,snakes,automata,backgrounds,fit,petri,bayesnet}
\usepackage{adjustbox}
\usepackage{hyperref}
\usepackage{url}
\usepackage{algorithmic}
\usepackage{algorithm}
\usepackage{mathtools}
\usepackage{amsmath}
\usepackage{amssymb}
\usepackage{wrapfig}
\usepackage{bm}
\usepackage{subcaption}
\usepackage{psfrag}
\usepackage{tikz}
\usepackage{epstopdf}

\newcommand{\beq}{\vspace{0mm}\begin{equation}}
	\newcommand{\eeq}{\vspace{0mm}\end{equation}}
\newcommand{\beqs}{\vspace{0mm}\begin{eqnarray}}
	\newcommand{\eeqs}{\vspace{0mm}\end{eqnarray}}
\newcommand{\barr}{\begin{array}}
	\newcommand{\earr}{\end{array}}
\newcommand{\Amat}[0]{{{\bf A}}}
\newcommand{\Bmat}{{\bf B}}
\newcommand{\Cmat}{{\bf C}}
\newcommand{\Dmat}{{\bf D}}
\newcommand{\Emat}[0]{{{\bf E}}}

\newcommand{\Hmat}{{\bf H}}
\newcommand{\Imat}{{\bf I}}

\newcommand{\Umat}[0]{{{\bf U}}}
\newcommand{\Vmat}[0]{{{\bf V}}}
\newcommand{\Wmat}[0]{{{\bf W}}}

\newcommand{\Ymat}{{\bf Y}}
\newcommand{\Zmat}{{\bf Z}}

\newcommand{\av}[0]{{\boldsymbol{a}}}
\newcommand{\bv}[0]{{\boldsymbol{b}}}
\newcommand{\cv}[0]{{\boldsymbol{c}}}
\newcommand{\dv}{\boldsymbol{d}}
\newcommand{\ev}[0]{{\boldsymbol{e}}}

\newcommand{\hv}[0]{{\boldsymbol{h}}}

\newcommand{\vv}{\boldsymbol{v}}
\newcommand{\wv}{\boldsymbol{w}}
\newcommand{\xv}{\boldsymbol{x}}
\newcommand{\yv}{\boldsymbol{y}}
\newcommand{\zv}{\boldsymbol{z}}

\newcommand{\etav}[0]{{\boldsymbol{\eta}}}

\newcommand{\thetav}[0]{\boldsymbol{\theta}}

\newcommand{\lambdav}[0]{{\boldsymbol{\lambda}}}
\newcommand{\muv}[0]{{\boldsymbol{\mu}}}

\newcommand{\xiv}[0]{{\boldsymbol{\xi}}}

\newcommand{\piv}{\boldsymbol{\pi}}
\newcommand{\rhov}[0]{{\boldsymbol{\rho}}}
\newcommand{\sigmav}[0]{{\boldsymbol{\sigma}}}

\newcommand{\phiv}{\boldsymbol{\phi}}
\newcommand{\chiv}[0]{{\boldsymbol{\chi}}}

\newcommand{\R}{\mathbb{R}}
\newcommand{\E}{\mathbb{E}}

\newcommand{\Lcal}{\mathcal{L}}
\newcommand{\Ncal}{\mathcal{N}}

\icmltitlerunning{Factored Temporal Sigmoid Belief Networks for Sequence Learning}

\begin{document}
\twocolumn[
\icmltitle{Factored Temporal Sigmoid Belief Networks for Sequence Learning}

\icmlauthor{Jiaming Song$^\dag$}{jiaming.tsong@gmail.com}
\icmlauthor{Zhe Gan$^\ddag$}{zhe.gan@duke.edu}
\icmlauthor{Lawrence Carin$^\ddag$}{lcarin@duke.edu}
\icmladdress{$^\dag$Department of Computer Science and Technology, Tsinghua University, Beijing, 100084, China}
\vspace{-0.25cm}\icmladdress{$^\ddag$Department of Electrical and Computer Engineering, Duke
	University, Durham, NC 27708, USA}

\icmlkeywords{}

\vskip 0.3in
]

\begin{abstract}
	Deep conditional generative models are developed to simultaneously learn the temporal dependencies of multiple sequences. The model is designed by introducing a three-way weight tensor to capture the multiplicative interactions between \textit{side information} and sequences. The proposed model builds on the Temporal Sigmoid Belief Network (TSBN), a sequential stack of Sigmoid Belief Networks (SBNs). The transition matrices are further factored to reduce the number of parameters and improve generalization. When side information is not available, a general framework for semi-supervised learning based on the proposed model is constituted, allowing robust sequence classification. Experimental results show that the proposed approach achieves state-of-the-art predictive and classification performance on sequential data, and has the capacity to synthesize sequences, with controlled style transitioning and blending. 
\end{abstract}

\section{Introduction}
The Restricted Boltzmann Machine (RBM) is a well-known undirected generative model with state-of-the-art performance on various problems. It serves as a building block for many deep generative models, such as the Deep Belief Network (DBN) ~\cite{hinton2006fast} and the Deep Boltzmann Machine (DBM) ~\cite{salakhutdinov2009deep}. Other variants of the RBM have been used for modeling discrete time series data, such as human motion \cite{taylor2006modeling}, videos \cite{sutskever2009recurrent}, and weather prediction \cite{mittelman2014structured}. Among these variants is the Conditional Restricted Boltzmann Machine (CRBM) \cite{taylor2006modeling}, where the hidden and visible states at the current time step are dependent on directed connections from observations at the last few time steps. Most RBM-based models use Contrastive Divergence (CD) \cite{hinton2002training} for efficient learning.

Recently, there has been a surging interest in deep \textit{directed} generative models, with applications in both static and dynamic data. In particular, advances in variational methods \cite{mnih2014neural,kingma2013auto,rezende2014stochastic} have yielded scalable and efficient learning and inference for such models, avoiding poor inference speed caused by the ``explaining away'' effect \cite{hinton2006fast}. One directed graphical model that is related to the RBM is the Sigmoid Belief Network (SBN) \cite{neal1992connectionist}.

The Temporal Sigmoid Belief Network (TSBN)\cite{TSBN_NIPS2015} is a fully directed generative model for discretely sampled time-series data, defined by a sequential stack of SBNs. The hidden state for each SBN is inherited from the states of the previous SBNs in the sequence. The TSBN can be regarded as a generalization of Hidden Markov Models (HMM), with compact hidden state representations, or as a generalization of Linear Dynamical Systems (LDS), characterized by non-linear dynamics. The model can be utilized to analyze many kinds of data, \textit{e.g.}, binary, real-valued, and counts, and has demonstrated state-of-the-art performance in many tasks.

However, the TSBN exhibits certain limitations, in that it does not discriminate different types of sequences when training, nor does it utilize (often available) side information during generation. For example, in the case of modeling human motion, the TSBN does not regard ``walking'' and ``running'' as different motions, and the style of generated motions is not controlled. To allow for conditional generation, we first propose a straightforward modification of the TSBN, introducing three-way tensors for weight transitions, from which weight matrices for the TSBN are extracted according to the side information provided. We then adopt ideas from \citet{taylor2009factored}, where the three-way weight tensor is factored into the multiplication of matrices, effectively reducing the model parameters. 
In our case, factoring is not directly imposed on the tensor parameters but over the style-dependent transition matrices, which provides a more-compact representation of the transitions, where sequences with different attributes are modeled by a shared set of parameters.
The model is able to capture the subtle similarities across these sequences, while still preserving discriminative attributes. Experiments show that the factored model yields better prediction performance than its non-factored counterpart.


For cases where side information is not available for the entire training set, we propose a framework for semi-supervised learning, inspired by \citet{kingma2014semi}, to infer that information from observations (\textit{e.g.}, by using a classifier to infer the motion style), while simultaneously training a generative model. 

The principal contributions of this paper are as follows: (\emph{i}) A fully \emph{directed} deep generative model for sequential data is developed, that permits fast conditional generation of synthetic data, and controlled transitioning and blending of different styles. (\emph{ii}) A new factoring method is proposed, that reduces the number of parameters and improves performance across multiple sequences. (\emph{iii}) A general framework for semi-supervised learning is constituted, allowing robust sequence classification. (\emph{iv}) A new recognition model with factored parameters is utilized to perform scalable learning and inference.

\section{Model Formulation}
%
\subsection{Temporal Sigmoid Belief Network}
The Temporal Sigmoid Belief Network (TSBN) \cite{TSBN_NIPS2015} models uniformly sampled time series data of length $T$ through a sequence of SBNs, such that at any given time step SBN biases depend on the states of SBNs in the previous time steps. Assume that each observation is real-valued, and the $t$-th time step is denoted $\vv_t \in \R^M$. The TSBN defines the joint probability of visible data $\Vmat$ and hidden states $\Hmat$ as
\begin{align}
	p_{\thetav}(\Vmat, \Hmat) = \prod_{t=1}^{T}p(\hv_t|\hv_{t-1}, \vv_{t-1}) \cdot p(\vv_t|\hv_{t},\vv_{t-1}) \label{eq:tsbn}
\end{align}
where $\Vmat = [\vv_1, \ldots, \vv_T]$, $\Hmat = [\hv_1, \ldots, \hv_T]$, and $\hv_t \in \{0, 1\}^J$ is the hidden state corresponding to time step $t$. Each conditional distribution in (\ref{eq:tsbn}) is expressed as
\begin{align}
	p(h_{jt}=1|\hv_{t-1}, \vv_{t-1})= \sigma(\tilde{h}_{jt}) \label{eq:tsbn_v_sigmoid}  \\
	p(\vv_t|\hv_t, \vv_{t-1}) = \Ncal(\muv_t, \mbox{diag}(\sigmav_t^2)) \\
	\tilde{\hv}_t = \Wmat_{1} \hv_{t-1} + \Wmat_{3} \vv_{t-1} + \bv \label{eq:tsbn_h}  
\end{align}
\begin{align}
	\muv_t = \Wmat_{2} \hv_t  + \Wmat_{4} \vv_{t-1} + \cv  \label{eq:tsbn_v}  \\
	\log\sigmav_t^2 = \Wmat_{2}^{\prime} \hv_t  + \Wmat_{4}^{\prime} \vv_{t-1} + \cv^{\prime}  \label{eq:tsbn_sigma}
\end{align}
where $\sigma(x) = 1 / (1 + \exp (-x))$, and $\mbox{diag}(\vv)$ is the diagonal matrix whose diagonal entries are $\vv$. $\hv_0$ and $\vv_0$ are defined as zero vectors. The model parameters, $\thetav$, are specified as $\Wmat_1 \in \R^{J \times J}$, $\Wmat_3 \in \R^{J \times M}$, $\{\Wmat_2, \Wmat_2^{\prime}\} \in \R^{M \times J}$, $\{\Wmat_4, \Wmat_4^{\prime}\} \in \R^{M \times M}$, $\bv \in \R^J$ and $\{\cv, \cv^{\prime}\} \in \R^M$.

The TSBN with $\Wmat_3 = \bm{0}$ and $\{\Wmat_4, \Wmat_4^{\prime}\} = \bm{0}$ is an HMM with an exponentially large state space. Specifically, in the HMM, each hidden state is represented as a one-hot-encoded vector of length $J$, while the hidden states of the TSBN are encoded by length-$J$ binary vectors, which can express $2^J$ different states. Compared with the Temporal Restricted Boltzmann Machine (TRBM) \cite{sutskever2007learning}, the TSBN is fully directed, allowing fast sequence generation from the inferred model.
\subsection{Conditional Temporal SBN}
\begin{figure}[t!]
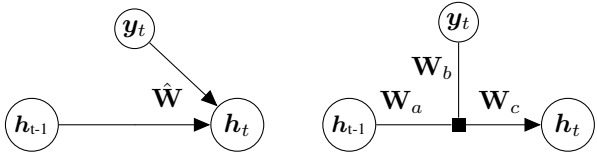

	\centering
	\begin{subfigure}[t]{0.2\textwidth}
		\centering
		\tikz {
			\node[latent, minimum size=20pt] (htm) {$\mbox{\footnotesize $\hv_{\text{t-1}}$}$};
			\node[right=of htm, rectangle, fill=black,minimum size=0pt, inner
			sep=0pt, node distance=1.0] (y) {};
			\node[latent, right=of y, minimum size=20pt] (ht) {$\hv_t$};
			\node[latent, above=of y, minimum size=15pt, node distance=0.6] (yt) {$\yv_t$};
			\draw (1.8,0.4) node {{$\hat{\Wmat}$}};
			\edge{htm}{ht}
			\edge{yt}{ht}
		}
	\end{subfigure}
	~ ~ ~ ~
	\begin{subfigure}[t]{0.2\textwidth}
		\centering
		\tikz {
			\node[latent, minimum size=20pt] (htm) {$\mbox{\footnotesize $\hv_{\text{t-1}}$}$};
			\node[right=of htm, rectangle, fill=black,minimum size=5pt, inner
			sep=0pt, node distance=1.0] (y) {};
			\node[latent, right=of y, minimum size=20pt] (ht) {$\hv_t$};
			\node[latent, above=of y, node distance=0.6, minimum size=15pt] (yt) {$\yv_t$};
			\factoredge{htm, yt}{y}{ht}
			\draw (1.1,0.75) node {{$\Wmat_b$}};
			\draw (0.7,0.3) node {{$\Wmat_a$}};
			\draw (2.0,0.3) node {{$\Wmat_c$}};
		}	
	\end{subfigure}
	\caption{Graphical illustration for non-factored weights (left) and factored weights (right) from $\hv_{t-1}$ to $\hv_{t}$.}
	\label{fig:factoring}
	\vspace{-2.9mm}
\end{figure}

We consider modeling \textit{multiple} styles of time series data by exploiting additional side information, which can appear in the form of ``one-hot'' encoded vectors in the case of style labels, or real-valued vectors in other situations (\textit{e.g.}, the longitude and latitude information in the weather prediction task considered in Section \ref{sec:weather_pred}). Let $\yv_t \in \R^S$ denote the side information at time step $t$. The joint probability is now described as \\
\begin{adjustbox}{minipage=1.29\linewidth,scale=0.9}
	\begin{align}
		p_{\thetav}(\Vmat, \Hmat|\Ymat) = \prod_{t=1}^{T}p(\hv_t|\hv_{t-1}, \vv_{t-1}, \yv_t)\cdot p(\vv_t|\hv_{t},\vv_{t-1}, \yv_t) \label{eq:ctsbn} \MoveEqLeft[3.15]
	\end{align}
\end{adjustbox} 
where $\Ymat = [\yv_1, \ldots, \yv_T]$. A straightforward method to incorporate side information is by modifying (\ref{eq:tsbn_h})-(\ref{eq:tsbn_sigma}) to allow the weight parameters to be dependent on $\yv_t$. Specifically,
\begin{align}
	\tilde{\hv}_t &= \Wmat_{1}^{(y)} \hv_{t-1} + \Wmat_{3}^{(y)} \vv_{t-1} + \bv^{(y)}  \label{eq:new_hath} \\
	\muv_t &= \Wmat_{2}^{(y)} \hv_t  + \Wmat_{4}^{(y)} \vv_{t-1} + \cv^{(y)}   \\
	\log\sigmav_t^2 &= \Wmat_{2}^{\prime(y)} \hv_t  + \Wmat_{4}^{\prime(y)} \vv_{t-1} + \cv^{\prime(y)}  \label{eq:new_sigmav}
\end{align}
where $\bv^{(y)} = \Bmat \yv_t$, $\cv^{(y)} = \Cmat \yv_t$; for $i \in \{1, 2, 3, 4\}$, $\Wmat_{ijk}^{(y)} = \sum_{s=1}^{S} \hat{\Wmat}_{ijks} y_{st}$, and $\hat{\Wmat}_i$ is a three-way tensor, shown in Figure \ref{fig:factoring} (left).\footnote{\small{$ \Wmat_{2}^{\prime(y)}$ and $ \Wmat_{4}^{\prime(y)}$ are also similarly defined, and we omit them when discussing $ \Wmat_{2}^{(y)}$ and $  \Wmat_{4}^{(y)}$ in the following sections. $\bv^{(y)}$, $\cv^{(y)}$ and $ \Wmat_{ijk}^{(y)}$ are dependent on $t$, but we choose to omit $t$ for simplicity.}} 
The model parameters are specified as $\hat{\Wmat}_1 \in \R^{J \times J \times S}$, $\hat{\Wmat}_2 \in \R^{M \times J \times S}$, $\hat{\Wmat}_3 \in \R^{J \times M \times S}$, $\hat{\Wmat}_4 \in \R^{M \times M \times S}$, $\Bmat \in \R^{J \times S}$, and $\Cmat \in \R^{M \times S}$. When the side information are one-hot encoded vectors, this is equivalent to training one TSBN for each style of sequences. We name this model the Conditional Temporal Sigmoid Belief Network (CTSBN).

\begin{figure}[t!]
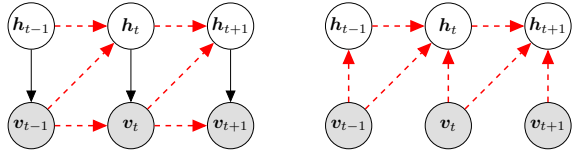

	\centering
	\begin{subfigure}[t]{0.2\textwidth}
		\centering
		\scalebox{0.7}{
			\tikz{
				\node[obs, minimum size=25pt] (vtm) {$\vv_{t-1}$};
				\node[obs, right=of vtm, minimum size=25pt] (vt)  {$\vv_t$};
				\node[obs, right=of vt, minimum size=25pt] (vtp) {$\vv_{t+1}$};
				\node[latent, above=of vtm, minimum size=25pt] (htm) {$\hv_{t-1}$};
				\node[latent, above=of vt, minimum size=25pt] (ht) {$\hv_t$};
				\node[latent, above=of vtp, minimum size=25pt] (htp) {$\hv_{t+1}$};
				\edge {htm} {vtm}
				\edge [red, thick, dashed]{htm, vtm} {ht}
				\edge [red, thick, dashed] {vtm} {vt}
				\edge [red, thick, dashed] {vt} {vtp}
				\edge {ht} {vt}
				\edge [red, thick, dashed]{ht, vt} {htp}
				\edge {htp} {vtp}
			}	
		}
		
	\end{subfigure}
	~ ~ ~ ~
	\begin{subfigure}[t]{0.2\textwidth}
		\centering
		\scalebox{0.7}{
			\tikz{
				\node[obs, minimum size=25pt] (vtm) {$\vv_{t-1}$};
				\node[obs, right=of vtm, minimum size=25pt] (vt)  {$\vv_t$};
				\node[obs, right=of vt, minimum size=25pt] (vtp) {$\vv_{t+1}$};
				\node[latent, above=of vtm, minimum size=25pt] (htm) {$\hv_{t-1}$};
				\node[latent, above=of vt, minimum size=25pt] (ht) {$\hv_t$};
				\node[latent, above=of vtp, minimum size=25pt] (htp) {$\hv_{t+1}$};
				\edge[red, thick, dashed]{vtm} {htm}
				\edge[red, thick, dashed]{vtm, htm, vt} {ht}
				\edge[red, thick, dashed]{vt, ht, vtp} {htp}
			}
		}
	\end{subfigure}
	\caption{Generative model (left) and recognition model (right) of the FCTSBN. Red, dashed arrows indicate factorized weights described in Figure \ref{fig:factoring} (right).}
	\label{fig:gmodel}
	\vspace{-3.0mm}
\end{figure}
\subsection{Factoring Weight Parameters}
While the CTSBN enables conditional generation, it has several disadvantages: (\textit{i}) the number of parameters is proportional to $S$ (recall that this is the number of different styles for the time-series data), which is prohibitive for large $S$; (\textit{ii}) parameters for different attributes/styles are not shared (no sharing of ``statistical strength''), therefore the model fails to capture the underlying regularities between different data, resulting in poor generalization.

To remedy these drawbacks, we factor the weight matrices $\Wmat_i^{(y)}$ defined in (\ref{eq:new_hath})-(\ref{eq:new_sigmav}), for $i \in \{1, 2, 3, 4\}$, as
\begin{align}
	\Wmat^{(y)} = \Wmat_a \cdot \textrm{diag}(\Wmat_b \yv_t) \cdot \Wmat_c  \label{eq:factor}
\end{align}
Suppose $\Wmat^{(y)} \in \R^{J \times M}$, then $\Wmat_a \in \R^{J \times F}$, $\Wmat_b \in \R^{F \times S}$ and $\Wmat_c \in \R^{F \times M}$, where $F$ is the number of factors. $\Wmat_a$ and $\Wmat_c$ are shared among different styles, which capture the input-to-factor and factor-to-output relationships, respectively; the diagonal term, $\textrm{diag}(\Wmat_b \yv_t)$, models the unique factor-to-factor relationship for each style. If the number of factors is comparable to that of other terms (\emph{i.e.}, $J$ and $M$), this reduces the number of parameters needed for all the $\Wmat^{(y)}$ from $J \cdot M \cdot S$ to $(J + M + S) \cdot F$, or equivalently, $O(N^3)$ to $O(N^2)$. In practice, we factor all parameters except for $\Wmat_2$ to improve generation by allowing a more flexible transition from $\hv_t$ to $\vv_t$ across multiple styles. \footnote{$\Wmat_2$ is independent to the order $n$ introduced in Section \ref{sec:experiments}; when $n$ is large, the number of parameters in $\Wmat_2$ is relatively small, thus factoring $\Wmat_2$ does not provide much advantage for reducing parameters.} We call this model the Factored Conditional Temporal Sigmoid Belief Network (FCTSBN).
\begin{figure}[t!]
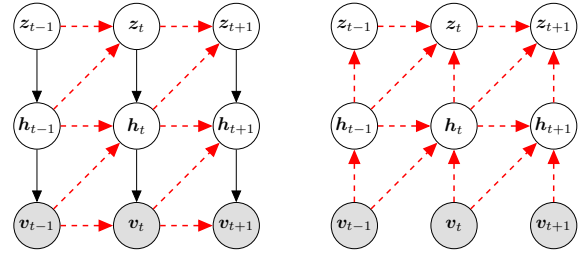

	\centering
	\begin{subfigure}[t]{0.2\textwidth}
		\centering
		\scalebox{0.7}{
			\tikz{
				\node[obs, minimum size=25pt] (vtm) {$\vv_{t-1}$};
				\node[obs, right=of vtm, minimum size=25pt] (vt)  {$\vv_t$};
				\node[obs, right=of vt, minimum size=25pt] (vtp) {$\vv_{t+1}$};
				\node[latent, above=of vtm, minimum size=25pt] (htm) {$\hv_{t-1}$};
				\node[latent, above=of vt, minimum size=25pt] (ht) {$\hv_t$};
				\node[latent, above=of vtp, minimum size=25pt] (htp) {$\hv_{t+1}$};
				\node[latent, above=of htm, minimum size=25pt] (ztm) {$\zv_{t-1}$};
				\node[latent, above=of ht, minimum size=25pt] (zt) {$\zv_{t}$};
				\node[latent, above=of htp, minimum size=25pt] (ztp) {$\zv_{t+1}$};
				\edge {ztm} {htm}
				\edge {zt} {ht}
				\edge {ztp} {htp}
				\edge [red, thick, dashed] {ztm, htm} {zt}
				\edge [red, thick, dashed] {zt, ht} {ztp}
				\edge {htm} {vtm}
				\edge [red, thick, dashed]{htm, vtm} {ht}
				\edge [red, thick, dashed] {vtm} {vt}
				\edge [red, thick, dashed] {vt} {vtp}
				\edge {ht} {vt}
				\edge [red, thick, dashed]{ht, vt} {htp}
				\edge {htp} {vtp}
			}	
		}
		
	\end{subfigure}
	~ ~ ~ ~
	\begin{subfigure}[t]{0.2\textwidth}
		\centering
		\scalebox{0.7}{
			\tikz{
				\node[obs, minimum size=25pt] (vtm) {$\vv_{t-1}$};
				\node[obs, right=of vtm, minimum size=25pt] (vt)  {$\vv_t$};
				\node[obs, right=of vt, minimum size=25pt] (vtp) {$\vv_{t+1}$};
				\node[latent, above=of vtm, minimum size=25pt] (htm) {$\hv_{t-1}$};
				\node[latent, above=of vt, minimum size=25pt] (ht) {$\hv_t$};
				\node[latent, above=of vtp, minimum size=25pt] (htp) {$\hv_{t+1}$};
				\node[latent, above=of htm, minimum size=25pt] (ztm) {$\zv_{t-1}$};
				\node[latent, above=of ht, minimum size=25pt] (zt) {$\zv_{t}$};
				\node[latent, above=of htp, minimum size=25pt] (ztp) {$\zv_{t+1}$};
				\edge[red, thick, dashed]{htm} {ztm}
				\edge[red, thick, dashed]{htm, ztm, ht} {zt}
				\edge[red, thick, dashed]{ht, zt, htp} {ztp}
				\edge[red, thick, dashed]{vtm} {htm}
				\edge[red, thick, dashed]{vtm, htm, vt} {ht}
				\edge[red, thick, dashed]{vt, ht, vtp} {htp}
			}
		}
	\end{subfigure}
	\caption{Generative model (left) and recognition model (right) of a deep FCTSBN with two layers. Red, dashed arrows indicate factorized weights described in Figure \ref{fig:factoring} (right).}
	\label{fig:dmodel}
	\vspace{-3.0mm}
\end{figure}
\subsection{Deep Architecture with FCTSBN}
The shallow model described above may be restrictive in terms of representational power. Therefore, we propose a deep architecture by adding stochastic hidden layers. We consider a deep FCTSBN with hidden layers $\hv_t^{(\ell)}$ for $t = 1 \ldots T$ and $\ell = 1 \ldots L$, where we denote $\hv_t^{(0)} = \vv_t$ and $\hv_t^{(L+1)} = \bm{0}$ for conciseness. Each of the hidden layers $\hv_t^{(\ell)}$ contains stochastic binary hidden variables, which is generated by $p(\hv_t^{(\ell)}) = \prod_{j=1}^{J^{(\ell)}} p(h_{jt}^{(\ell)}| \hv_t^{(\ell+1)}, \hv_{t-1}^{(\ell)}, \hv_{t-1}^{(\ell-1)}, \yv_t)$, where each conditional distribution is parameterized by a sigmoid function, as in (\ref{eq:tsbn_v_sigmoid}). Learning and inference for the model with stochastic hidden layers is provided in Section \ref{sec:inference_deep}. We choose not to consider deterministic layers here, as was considered in \citet{TSBN_NIPS2015}, since such complicates the gradient computation in the FCTSBN, while having similar empirical results compared with the usage of stochastic layers (we tried both in practice).
\subsection{Semi-supervised Learning with FCTSBN}
As expressed thus far, training the FCTSBN requires side information for each time step. In many applications \cite{le2014distributed,srivastava2015unsupervised}, however, unlabeled sequential data might be abundant, while obtaining the side information for all the data is expensive or sometimes impossible. We propose a framework for semi-supervised learning based on FCTSBN, with the capacity to simutaneously train a generative model and a classifier from labeled and unlabeled data.

Specifically, assume that the side information is generated from the prior distribution $p_{\thetav} (\Ymat; \piv)$ parametrized by $\piv$ (which can be a multinomial distribution if $\yv_t$ are labels), then the generative model can be described as
\begin{align}
	p_{\thetav} (\Vmat, \Hmat, \Ymat) = p_{\thetav} (\Ymat; \piv) \cdot p_{\thetav}(\Vmat, \Hmat | \Ymat)  \label{eq:semi_model}
\end{align}
where $p_{\thetav} (\Vmat, \Hmat | \Ymat)$ is the conditional generative model described in (\ref{eq:ctsbn}). Details on training $p_{\thetav}(\Vmat, \Hmat, \Ymat)$ are discussed in Section \ref{sec:semi_inference}.

\vspace{-5pt}
\section{Scalable Learning and Inference}
The exact posterior over the hidden variables in (\ref{eq:ctsbn}) is intractable, and methods like Gibbs sampling and mean-field variational Bayes (VB) inference, can be inefficient (particularly for inference at test time). We follow ~\citet{TSBN_NIPS2015}, and apply the Neural Variational Inference and Learning (NVIL) algorithm described in ~\citet{mnih2014neural}, allowing tractable and scalable parameter learning and inference by introducing a new recognition model.
\vspace{-5pt}
\subsection{Lower Bound Objective} \label{sec:recognition_model}
We are interested in training the CTSBN and FCTSBN models, $p_{\thetav}(\Vmat, \Hmat|\Ymat)$, both of which come in the form of (\ref{eq:ctsbn}). Given an observation $\Vmat$, we introduce a fixed-form distribution $q_{\phiv}(\Hmat|\Vmat,\Ymat)$ with parameters $\phiv$, to approximate the true posterior $p(\Hmat | \Vmat,\Ymat)$. According to the variational principle ~\cite{jordan1999an}, we construct a lower bound on the marginal log-likelihood with the following form \\
\begin{adjustbox}{minipage=1.05\linewidth,scale=0.95}
	\begin{align}
		\Lcal (\Vmat| \Ymat, \thetav, \phiv) = \mathcal{J}(q_{\phiv}(\Hmat|\Vmat, \Ymat) , p_{\thetav}(\Vmat, \Hmat| \Ymat))  \label{eq:lower_bound}
	\end{align}
\end{adjustbox}

where $\mathcal{J}(q, p) = \E_q[\log p - \log q]$ is used to save space, and $q_{\phiv}(\Hmat|\Vmat, \Ymat)$ is the recognition model. For both models, the recognition model is expressed as
\begin{align}
	q_{\phiv}(\Hmat | \Vmat, \Ymat) = \prod_{t=1}^{T} q(\hv_t| \hv_{t-1}, \vv_{t}, \vv_{t-1}, \yv_t)  \label{eq:recog}
\end{align}
and each conditional distribution is specified as
\begin{align}
	q(h_{jt} = 1|\hv_{t-1}, \vv_t, \vv_{t-1}, \yv_t) = \sigma(\hat{h}_{jt})  \\
	\hat{\hv}_t = \Umat_1^{(y)} \hv_{t-1} + \Umat_2^{(y)} \vv_{t} + \Umat_3^{(y)} \vv_{t-1} + \dv^{(y)} 
\end{align}
For CTSBN, $\Umat_{ijk}^{(y)} = \sum_{l=1}^{L} \hat{\Umat}_{ijkl} y_{lt}$, where $\hat{\Umat}_i$ is a three-way tensor similar to $\hat{\Wmat}_i$; for FCTSBN, the weight matrix $\Umat_i^{(y)}$ is factored as in (\ref{eq:factor}). The recognition model introduced in (\ref{eq:recog}), allows fast and scalable inference.
%
%
\subsection{Parameter Learning} \label{sec:inference_parameters}
To optimize the lower bound described in (\ref{eq:lower_bound}), we apply Monte Carlo integration to approximate the expectations over $q_{\phiv}$, and stochastic gradient descent (SGD) for parameter optimization. The gradients of $\Lcal$ with respect to $\thetav$ and $\phiv$ are expressed as \\
\begin{adjustbox}{minipage=1.05\linewidth,scale=0.97}
	\begin{align}
		\nabla_{\thetav} \Lcal &= \E_{q_{\phiv}(\Hmat|\Vmat, \Ymat)}[\nabla_{\thetav} \log p_{\thetav} (\Vmat, \Hmat | \Ymat)]  \label{eq:grad_theta} \\
		\nabla_{\phiv} \Lcal &= \E_{q_{\phiv}(\Hmat | \Vmat, \Ymat)}[l_{\phiv}(\Vmat, \Hmat, \Ymat) \cdot \nabla q_{\phiv}(\Hmat | \Vmat, \Ymat)] 
	\end{align}
\end{adjustbox}

where $l_{\phiv}=\log p_{\thetav}(\Vmat, \Hmat | \Ymat)-\log q_{\phiv} (\Hmat|\Vmat,\Ymat)$, termed the \textit{learning signal} for the recognition parameters $\phiv$. 

The expectation of $l_{\phiv}$ is exactly the lower bound (\ref{eq:lower_bound}). The recognition model gradient estimator, however, can be very noisy, since the estimated learning signal is potentially large. According to ~\citet{mnih2014neural}, we apply two variance-reduction techniques. The first technique is applied by centering the learning signal by substracting the data-dependent baseline and data-independent baseline learned during training. The second technique is variance normalization, which is normalizing the learning signal by a running estimate of its standard deviation. All learning details and evaluation of gradients are provided in Supplementary Sections B.1-B.3. 
%
%

\vspace{-1.0mm}
\subsection{Extension for deep models} \label{sec:inference_deep}
The recognition model with respect to the deep FCTSBN is shown in Figure \ref{fig:dmodel} (right). Since the middle layers are also stochastic, the calculation of the lower bound includes more terms, in the form of
\begin{align}
	\Lcal = \sum_{\ell=0}^{L} \mathcal{J} (q_{\phiv}(\Hmat^{(\ell+1)}| \Hmat^{(\ell)}, \Ymat) , p_{\thetav}(\Hmat^{(\ell+1)}, \Hmat^{(\ell)}| \Ymat))  \nonumber
\end{align}
Each layer has a unique set of parameters whose gradient is zero for the lower bound of the other layers. Therefore, we can calculate the gradients separately for each layer of parameters, in a similar fashion to the single-layer FCTSBN. All details are provided in Supplementary Section B.4.

\vspace{-1.0mm}
\subsection{Extension for Semi-supervised Learning} \label{sec:semi_inference}
The recognition model in Section \ref{sec:recognition_model} provides a fast bottom-up probabilistic mapping from observations to hidden variables. Following ~\citet{kingma2014semi}, for the semi-supervised model (\ref{eq:semi_model}), we introduce a recognition model for both $\Hmat$ and $\Ymat$, with the factorized form $q_{\phiv}(\Hmat,\Ymat|\Vmat) = q_{\phiv}(\Hmat|\Vmat,\Ymat) \cdot q_{\phiv}(\Ymat|\Vmat)$. The first term, $q_{\phiv}(\Hmat|\Vmat,\Ymat)$, is the same as in (\ref{eq:recog}). When $\yv_t$ is a one-hot encoded vector, we assume a discriminative classifier $q_{\phiv}(\yv_t |\vv_t) = \mathrm{Cat}(\vv_t; \pi_{\phiv}(\vv_t))$, where $\mathrm{Cat}(\xv; \piv)$ denotes the categorical distribution. 

When the label corresponding to a sequence is missing, it is treated as a latent variable over which we perform posterior inference. The variational lower bound for the unlabeled data is
\begin{align}
	\Lcal_u = \mathcal{J}(q_{\phiv}(\Hmat, \Ymat|\Vmat) , p_{\thetav} (\Hmat, \Vmat, \Ymat))  \label{eq:lower_bound_unlabeled}
\end{align}
When the label is observed, the variational bound is a simple extension of (\ref{eq:lower_bound}), expressed as \\
\vspace{-1.0mm}
\begin{adjustbox}{minipage=1.08\linewidth,scale=0.93}
	\begin{align}
		\Lcal = \mathcal{J}(q_{\phiv}(\Hmat|\Vmat, \Ymat) , p_{\thetav}(\Vmat, \Hmat| \Ymat)) + \log p_{\thetav}(\Ymat;\rhov)  \label{eq:bound_labeled}
	\end{align}
\end{adjustbox}

where $p_{\thetav}(\Ymat;\rhov)$ is the prior distribution for $\Ymat$, a constant term that can be omitted. To exploit the discriminative power of the labeled data, we further add a classification loss to (\ref{eq:bound_labeled}). The resulting objective function is
\begin{align}
	\Lcal_l = \Lcal + \alpha \cdot \E_{\tilde{p}_l(\Vmat, \Ymat)}[\log q_{\thetav}(\Ymat|\Vmat)] 
\end{align}
where $\Lcal$ is the lower bound for the generative model, defined in (\ref{eq:bound_labeled}), and $\tilde{p}_{l}(\Vmat, \Ymat)$ denotes the empirical distribution. Parameter $\alpha$ controls the relative weight between the generative and discriminative components within the semi-supervised learning framework\footnote{We use $\alpha=2 \cdot T$ throughout the experiments, but obtain similar performance for other values of $\alpha$ within $[0.1 \cdot T, 4 \cdot T]$.}, where the extended objective function $\Lcal_s = \Lcal_l + \Lcal_u$ takes both labeled and unlabeled data into account. Details for optimizing $\Lcal_s$ are included in Supplementary Section B.5.

\section{Related Work}
Probabilistic models for sequential data in the deep learning literature can be roughly divided into two categories. The first category includes generative models \emph{without} latent variables, which rely on use of Recurrent Neural Networks (RNNs) \cite{sutskever2011generating,graves2013generating,chung2015gated}. The second category contains generative models \emph{with} latent variables, which can be further divided into two subcategories: (\emph{i}) \emph{undirected} latent variable models, utilizing the RBM as the building block \cite{taylor2006modeling,sutskever2007learning,sutskever2009recurrent,boulanger2012modeling,mittelman2014structured}; (\emph{ii}) \emph{directed} latent variable models, \emph{e.g.}, extending the variational auto-encoders \cite{bayer2014learning,fabius2014variational,chung2015recurrent}, or utilizing the SBN as the building block \cite{gan2015scalable,gan2015learning,TSBN_NIPS2015}. 


Among this work, the Factored Conditional Restricted Boltzmann Machine (FCRBM) \cite{taylor2009factored} and Temporal Sigmoid Belief Network (TSBN) \cite{TSBN_NIPS2015} are most related to the work reported here. However, there exist several key differences. Compared with FCRBM, our proposed model is fully \emph{directed}, allowing fast generation through ancestral sampling; while in the FCRBM, alternating Gibbs sampling is required to obtain a sample at each time-step. Compared with TSBN, where generation of different styles was purely based on initialization, our model utilizes side information to gate the connections of a TSBN, which makes the model context-sensitive, and permits controlled transitioning and blending (detailed when presenting experiments in Section \ref{sec:experiments}). 


Another novelty of our model lies in the utilization of a factored three-way tensor to model the multiplicative interactions between side information and sequences. Similar ideas have also been exploited in \cite{memisevic2007unsupervised,taylor2009factored,sutskever2011generating,kiros2014multimodal,kiros2014multiplicative}.  We note that our factoring method is different from the one used in FCRBM. Specifically, in FCRBM, the energy function was factored, while our model factors over the transition parameters, by which we capture the underlying similarities within different sequences, boosting performance.

Most deep generative models focus on exploring the generative ability \cite{kingma2013auto}, and little work has been done on examining the discriminative ability of deep generative models, except \citet{kingma2014semi,li2015max}. However, both works are restricted to the application of static data. Our paper is the first to develop a semi-supervised sequence classification method with deep generative models. 

In terms of inference, the wake-sleep algorithm \cite{hinton1995wake}, Stochastic Gradient Variational Bayes (SGVB) \cite{kingma2013auto} and Neural Variational Inference and Learning (NVIL) \cite{mnih2014neural} are widely studied for training recognition models. We utilize NVIL for scalable inference, and our method is novel in designing a new factored multiplicative recognition model.  




\section{Experiments} \label{sec:experiments}
Sections \ref{sec:Mocap2}-\ref{sec:transitions} report the results of training several models with data from the CMU Motion Capture Database. Specifically, we consider two datasets: (\emph{i}) motion sequences performed by subject 35 \cite{taylor2006modeling}(\textbf{mocap2}), which contains two types of motions, \emph{i.e.}, walking and running; (\emph{ii}) motion sequences performed by subject 137 \cite{taylor2009factored}(\textbf{mocap10}), which contains 10 different styles of walking. Both datasets are preprocessed using downsampling and scaling to have zero mean and unit variance. In Section \ref{sec:weather_pred} and \ref{sec:text_gene}, we present additional experiments on weather prediction and conditional text generation to further demonstrate the versatility of the proposed model. The weather prediction dataset \cite{liu2010learning} contains monthly observations of time series data of 18 climate agents (data types) over different places in North America.

The FCTSBN model with $\Wmat_3^{(y)} \equiv {\bf 0}$ and $\Wmat_4^{(y)} \equiv {\bf 0}$ is denoted Hidden Markov FCTSBN. The deep extension to FCTSBN is abbreviated as dFCTSBN. We use this abbreviation to denote the conditional generative model instead of the semi-supervised model, unless explicitly stated. Furthermore, we allow each observation to be dependent on the hidden and visible states of the previous $n$ time steps, instead of $n=1$. We refer to $n$ as the \emph{order} of the model.
The choice of $n$ can be different according to the specific scenario. For mocap2 prediction and mocap10 generation, $n$ is selected to align with previous methods on these tasks; for other prediction tasks, $n$ is chosen to balance between performance and model complexity; for semi-supervised tasks, $n=6$ is considered to allow consecutive frames of data to be utilized effectively by the classifier.

For CTSBN, the model parameters for weights are initialized by sampling from $\Ncal({\bf 0}, 0.001^2 \Imat)$, whereas the bias parameters are initialized as zero. For FCTSBN, the parameters are initialized differently, since the actual initialization value of the weight parameters depends on the product of factors. To ensure faster convergence, the initial values $\Wmat_a$, $\Wmat_b$ and $\Wmat_c$ are sampled from $\Ncal({\bf 0}, 0.01^2 \Imat)$. We use RMSprop \cite{tieleman2012lecture} throughout all the experiments. This is a stochastic gradient descent (SGD) method that allows the gradients to be adaptively rescaled by a running average of their recent magnitude. The data-dependent baseline is implemented with a single-hidden-layer neural network with 100 tanh units. We update the estimated learning signal with a momentum of 0.9.

The prediction of $\vv_t$ given $\vv_{1:t-1}$ requires first sampling $\hv_{1:t-1}$ from $q_{\phiv}(\hv_{1:t-1} | \vv_{1:t-1},\yv_{1:t-1})$, then calculating the posterior $p_{\thetav}(\hv_{t}| \hv_{1:t-1}, \vv_{1:t-1}, \yv_{t})$ for $\hv_t$. Given $\hv_t$, a prediction can be made for $\vv_t$ by sampling from $p_{\thetav}(\vv_t | \hv_{1:t}, \vv_{1:t-1}, \yv_t)$, or by using the expectation of sampled variables. 
Generating samples, on the other hand, is a similar but relatively easier task, where sequences can be generated through ancestral sampling. A special advantage of the conditional generative model appears when the side information $\yv_t$ changes over time, so that we can generate sequences with style transitions that do not appear in the training data. More details on generating such sequences are discussed in Section \ref{sec:transitions}.
\subsection{Mocap2 Prediction} \label{sec:Mocap2}
For the first experiment, we perform the prediction task using the mocap2 dataset. We used 33 running and walking sequences, partitioned them into 31 training sequences and 2 test sequences, as in ~\citet{TSBN_NIPS2015}. Our models have 100 hidden units (and 50 factors for factored models) in each layer and the order of $n=1$, according to the settings in ~\citet{TSBN_NIPS2015}. Deep models, including deep FCTSBN and deep TSBN, have two hidden layers of 100 units each. The side information, $\yv_t$, is a one-hot encoded vector of length 2, which indicates ``running'' or ``walking''.

From Table \ref{tab:mocap2_pred}, it is observed that the one-layered conditional models have a significant improvement over TSBN, ss(spike-slab)-SRTRBM ~\cite{mittelman2014structured} and g(Gaussian)-RTRBM ~\cite{sutskever2009recurrent}, whereas FCTSBN has better performance than CTSBN, due to the factoring mechanism. Results of deep FCTSBN have comparable performances to that of TSBN with stochastic hidden layers; this may be because only two styles of time-series data are considered. 
\begin{table}[t!]
	\centering
	\begin{tabular}{|c|c|c|}
		\hline
		Method & Walking & Running \\\hline
		FCTSBN & $\bm{4.59} \pm 0.35$ & $\bm{2.86} \pm 0.23$ \\
		CTSBN & $4.67 \pm 0.22$ & $3.41 \pm 0.65$ \\
		TSBN$^\circ$ & $5.12 \pm 0.50$ & $4.85 \pm 1.26$ \\
		\hline
		dFCTSBN & $\bm{4.31} \pm 0.13$ & $2.58 \pm 0.21$ \\
		DTSBN-S$^\circ$ & $4.40 \pm 0.28$ & $\bm{2.56} \pm 0.40$ \\
		DTSBN-D$^\circ$ & $4.62 \pm 0.01$ & $2.84 \pm 0.01$ \\
		\hline
		ss-SRTRBM$^\diamond$ & $8.13 \pm 0.06$ & $5.88 \pm 0.05$ \\
		g-RTRBM$^\diamond$ & $14.41 \pm 0.38$ & $10.91 \pm 0.27$ \\
		\hline
	\end{tabular}
	\caption{Prediction error obtained for the mocap2 dataset. ($^\circ$) taken from \citet{TSBN_NIPS2015}; ($^\diamond$) taken from \citet{mittelman2014structured}. Bold indicate the best results for shallow deep models, respectively.}
	\label{tab:mocap2_pred}
	\vspace{-3mm}
\end{table}
\subsection{Mocap10 Generation and Prediction}  \label{sec:Mocap10}
In order to demonstrate the conditional generative capacity of our model, we choose the mocap10 dataset, which contains 10 different walking styles, namely \textit{cat}, \textit{chicken}, \textit{dinosaur}, \textit{drunk}, \textit{gangly}, \textit{graceful}, \textit{normal}, \textit{old-man}, \textit{sexy} and \textit{strong}. A single-layer Hidden Markov FCTSBN with 100 hidden units and order of $n=12$ is trained on the dataset for 400 epochs, whereas the parameters are updated 10 times for each epoch. The side information $\yv_t$ is provided as a one-hot encoded vector of length 10. We set a fixed learning rate of $3 \times 10^{-3}$, and a decay rate of $0.9$.
Motions of a particular style are generated by initializing with 12 frames of the training data for that style and fixing the side information during generation.

Our one-layer model can generate walking sequences of 9 styles out of 10, which are included in the supplementary videos. The only exception is the \textit{old-man} style, where our model captures the general posture but fails to synthesize the subtle frame-to-frame changes in the footsteps.

We also present prediction results. For each style, we select 90\% of the sequences as training data, and use the rest as testing data. We consider FCTSBN and a two-layer deep FCTSBN with order $n=8$, 100 hidden variables and 50 factors on each layer. We also consider FCRBM with 600 hidden variables, 200 factors, 200 features and order of $n=12$, as in \citet{taylor2009factored}. The average prediction error over 10 styles for FCRBM, FCTSBN and dFCTSBN are $0.4355$, $0.4832$, and $0.3599$, respectively, but the performances for different styles vary significantly. Detailed prediction error results for each style is provided in Supplementary Section C.
\subsection{Semi-supervised learning with mocap2 and mocap10}  \label{sec:Semi}
\label{sec:exp_semi}
\begin{figure}[t!]
	\centering
	\includegraphics*[width=0.40\textwidth]{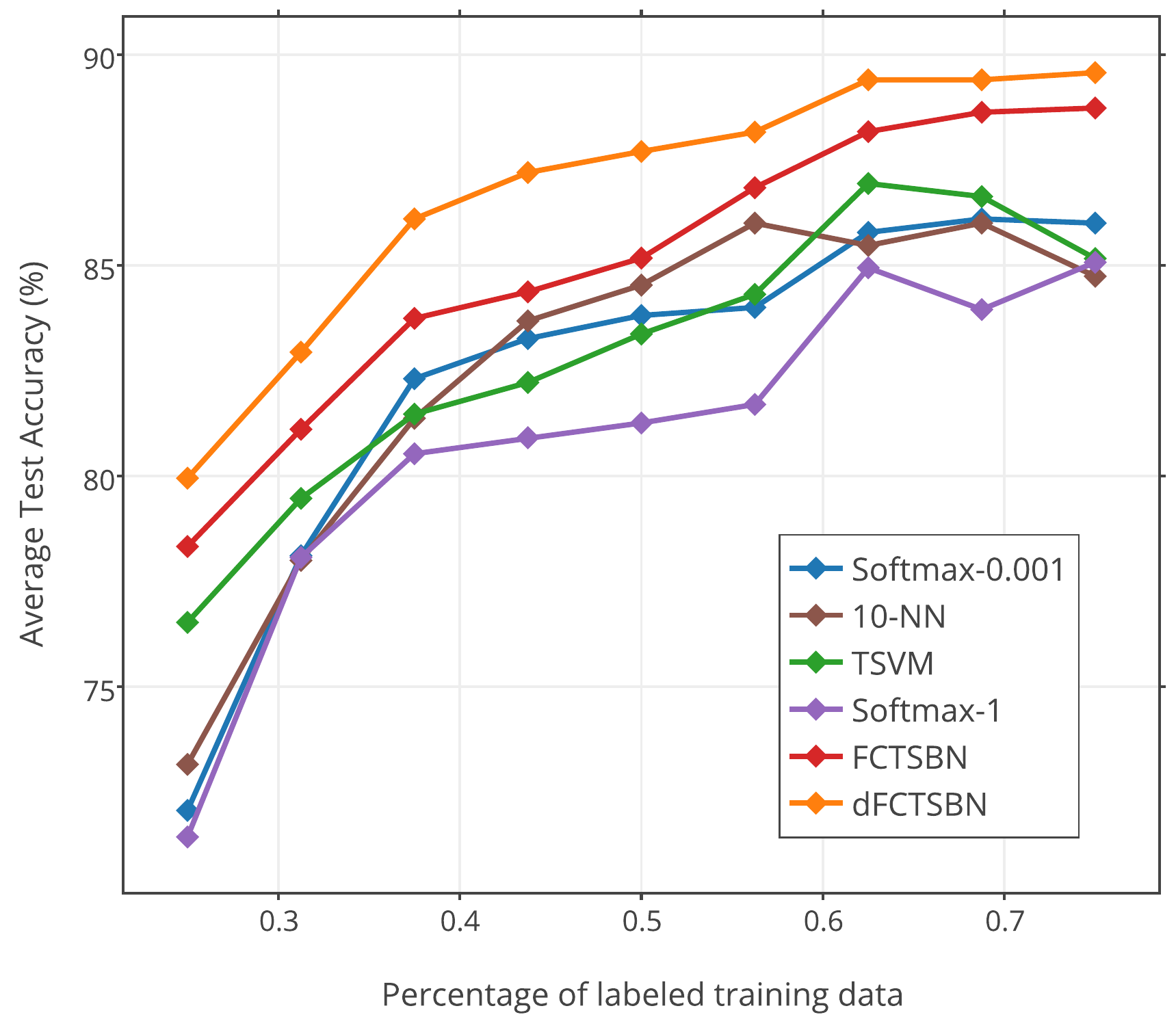}
	\caption{Test results on mocap10 classification. Each result is the average over 5 independent trials.}
	\label{fig:mocap10_classify}
	\vspace{-10pt}
\end{figure}
Our semi-supervised learning framework can train a classifier and a generative model simultaneously. In this section we present semi-supervised classification results for mocap10, and generated sequences for mocap2. \footnote{\small{We cannot perform both tasks on a single dataset due to the limitations of the mocap datasets: for mocap2, the 2-class classification task is so simple that the test accuracy is almost 1; for mocap10, the number of frames is too small to train a generative model with semi-supervised learning.}}

For mocap10, we perform 10-class classification over the motion style with windows of 7 consecutive frames from the sequences. We use FCTSBN and dFCTSBN of order $n=6$ with 100 hidden variables on each layer, set $q_{\phiv}(\Ymat|\Vmat)$ as a one-layer softmax classifier, and use $\alpha = 2 \cdot T$ throughout our experiments. Our baseline models include K-nearest neighbors (K-NN), the one-layer softmax classifier with regularization parameter $\lambda$ (Softmax-$\lambda$), and transductive support vector machines (TSVM) ~\cite{gammerman1998learning}. Sequences of certain styles are truncated such that each style has 500 frames of actual motions. We hold out 100 frames for testing, and the rest as training data, a consecutive proportion of which is provided with $\yv_t$.

Figure \ref{fig:mocap10_classify} shows the average test accuracy of different methods under various percentage of labeled training data, where our models consistently have better performance than other methods, even with a simple, one-layer classifier $q_{\phiv}$. To ensure fair comparison, the other parametric methods have roughly the same number of parameters. This would demonstrate that the $\Lcal_u$ term in (\ref{eq:lower_bound_unlabeled}) serves as an effective regularizer that helps prevent overfitting. Our deep FCTSBN model has better performance than the shallow model, which may be attributed to a more flexible regularizer due to more parameters in the generative model objective. 

For mocap2 generation, the training data are processed such that one sequence for each style is provided with $\yv_t$, whereas the remaining 31 sequences are unlabeled. Then the data are used to train a Hidden Markov FCTSBN of order $n=6$ with 100 hidden variables and 50 factors, and $q_{\phiv}(\Ymat | \Vmat)$ as a one-layer softmax classifier. Although the proportion of labeled data is very small, we are able to synthesize smooth, high-quality motion sequences for both styles. Related videos are presented in the Supplementary Material.
\subsection{Generating sequences with style transitions and blending}  \label{sec:transitions}
\begin{figure}[t!]
	\centering
	\includegraphics*[width=0.50\textwidth]{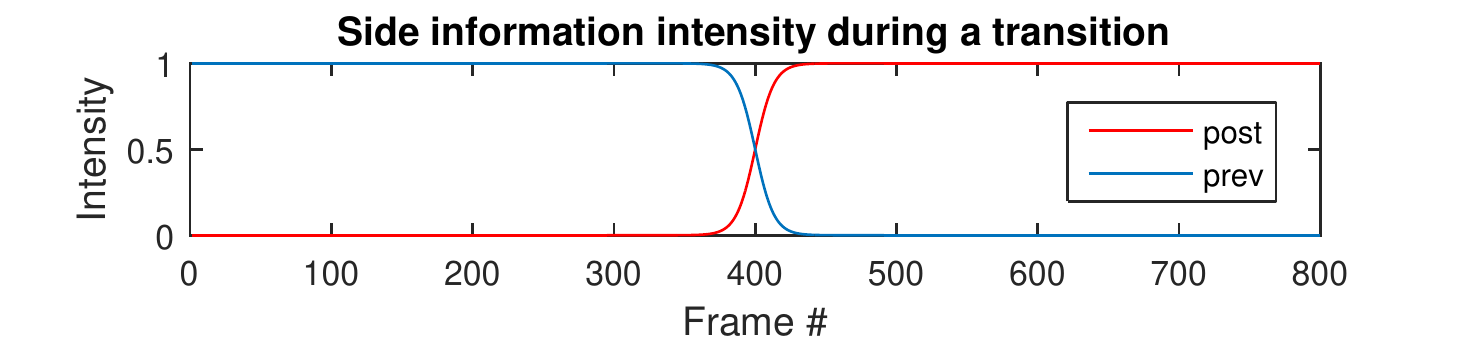}
	\caption{Intensity of $\yv_t$ for different styles during transition. Pre-transition style is denoted as ``prev''; post-transition style as ``post''. }
	\label{fig:transition}
	\vspace{-2.5mm}
\end{figure}

\begin{figure}[t!]
	\centering
	\begin{subfigure}{0.22\textwidth}
		\includegraphics*[width=\textwidth]{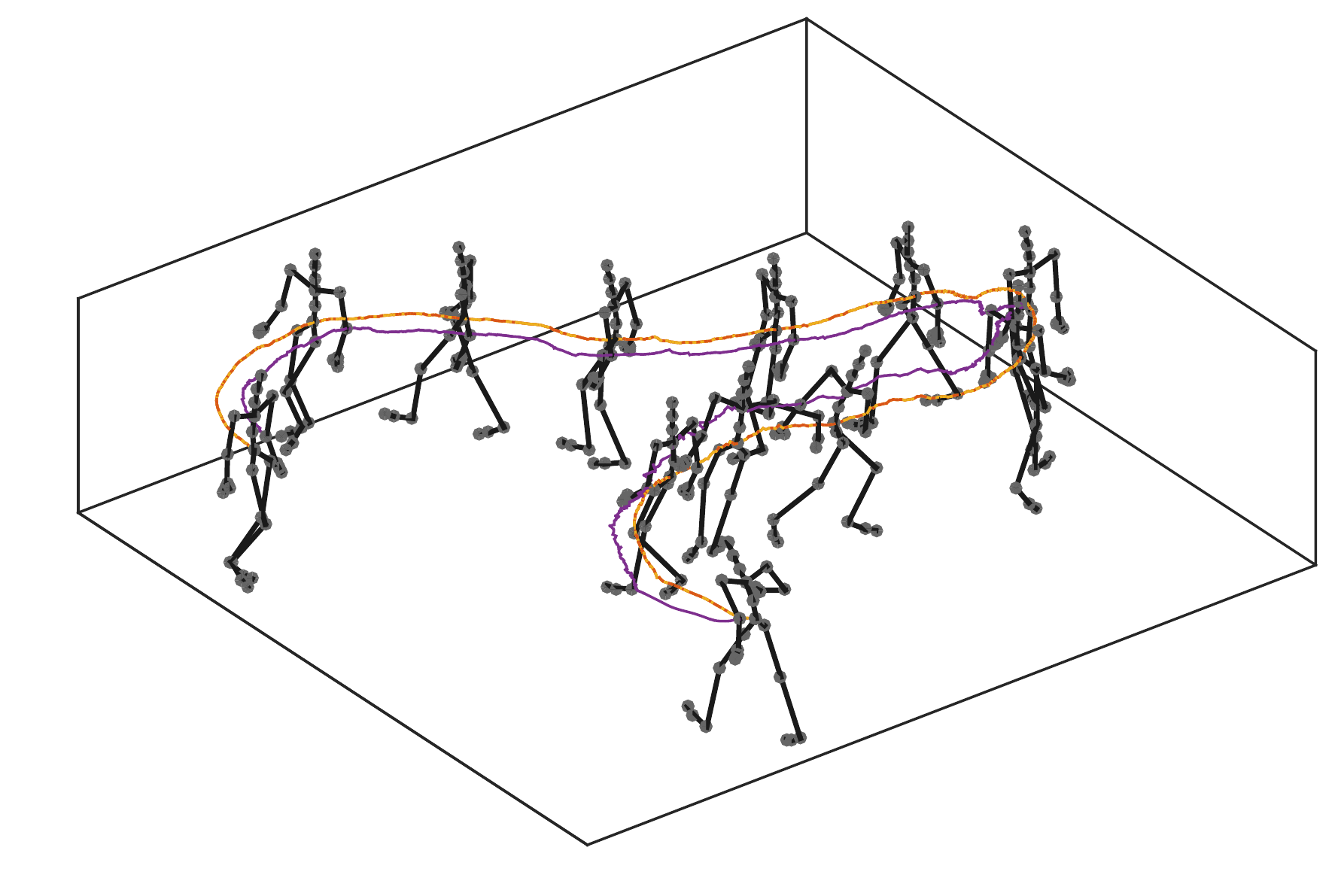}
	\end{subfigure}
	~~~~
	\begin{subfigure}{0.22\textwidth}
		\includegraphics*[width=\textwidth]{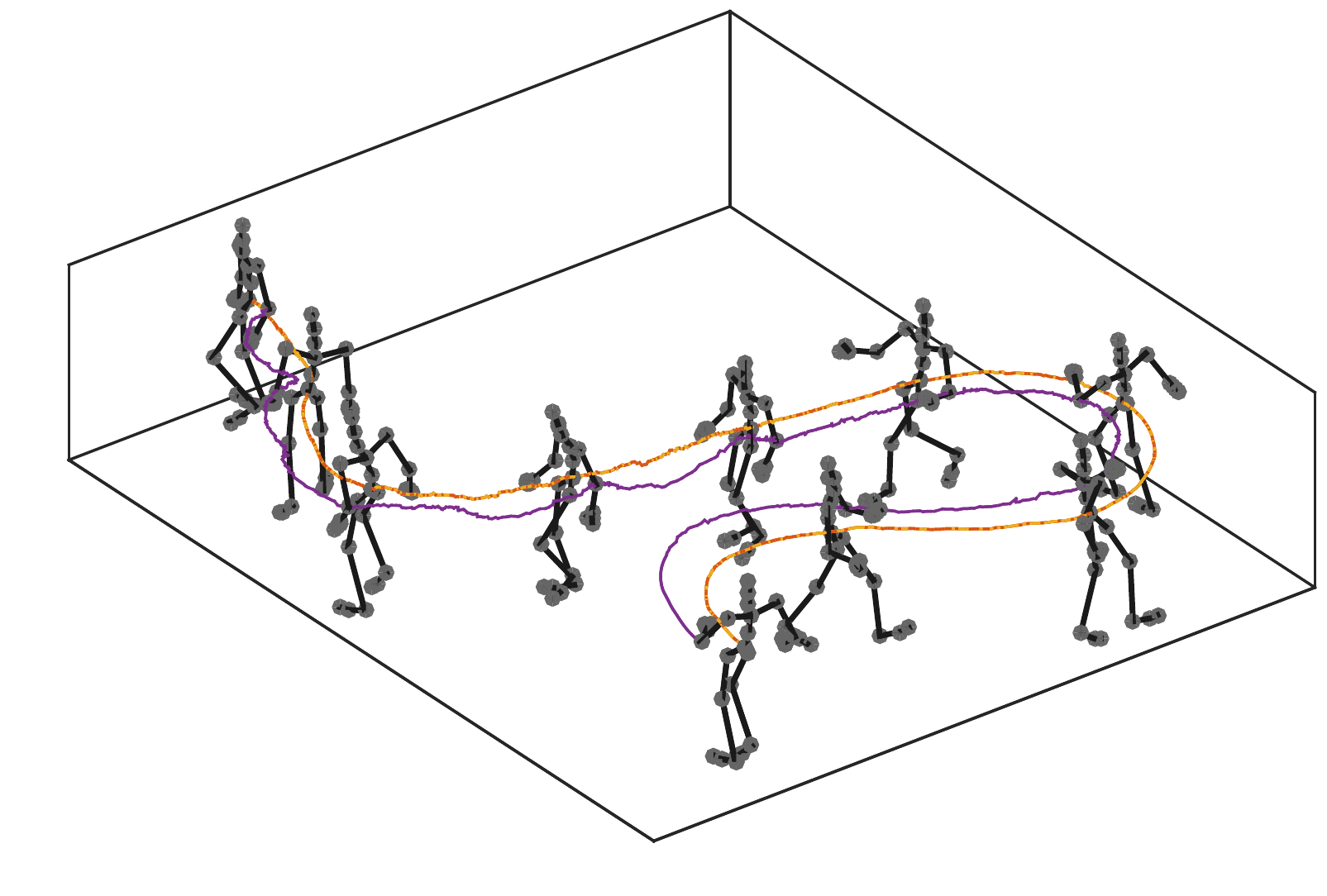}
	\end{subfigure}	
	\caption{\textbf{Left:} \textit{Drunk} to \textit{normal} (6 skeletons for each style). \textbf{Right:} \textit{Graceful} to \textit{gangly} (5 skeletons for each style). Both figures are generated within 800 frames, with the side information given according to Figure \ref{fig:transition}.}
	\label{fig:walking}
	\vspace{-10pt}
\end{figure}
One particular advantage of modeling multiple time-series data with one FCTSBN over training several TSBNs for each sequence is that the FCTSBN has a shared hidden representation for all styles, so $\yv_t$ are no longer restricted to a constant value during generation. This flexibility allows generation of sequences that do not appear in the training set, such as transitions between styles, or combinations of multiple styles.

Videos displaying smooth and efficient style transitions and blending are provided in the supplementary material. Two examples are provided in Figure \ref{fig:walking}. We also include transitions between walking and running with the generative model trained in Section \ref{sec:exp_semi}. To enable such transition, we transform the side information from one to another using a sigmoid-shaped form as in Figure \ref{fig:transition}. The entire transition process happens over around 60 frames, which is much faster than the 200-frame transition proposed for FCRBM ~\cite{taylor2009factored} \footnote{In practice, a sudden change of labels will result in "jerky" transitions~\cite{taylor2009factored}, but we aim to use as few frames as possible during transition.}. For style blending, the side information is provided by a convex combination of several one-hot-encoded side information vectors.
\subsection{Weather prediction}  \label{sec:weather_pred}
We demonstrate weather prediction in multiple places, and discuss how different representations of side information can affect performance. We select 25 places in the United States, which are located in a $5\times 5$ grid spanning across multiple climate regions. The weather data are processed so that for every month in 1990-2002, each place has 17 climate indicator values (\emph{e.g.}, temperature). We use only longitude and latitude as side information.

One way of providing $\yv_t$ is by using the geographic information directly, such that $\yv_t$ would be a real-valued vector of length 2 (we call this setting 2d for simplicity). 
Such representation is oversimplified, since the transition parameters $\Wmat$ of a certain location cannot be described as a linear combination of longitude and latitude. 
%
Therefore, we consider two other representations of side information. One uses a vector with 10 binary values (10d), which is the concatenation of two one-hot encoded vector of length 5 for longitude and latitude, respectively; the other assumes a one-hot encoded vector of 25, one for each location (25d).

We use the monthly climate data from 1990 to 2001 as training set, and the data of 2002 as test set. We consider CTSBN, FCTSBN, and two-layer deep FCTSBN of order $n=6$, with 100 hidden variables and 50 factors on each hidden layer. During test time, we predict the weather of every month in 2002 given the information of previous 6 months.
%
%
\begin{table}[t!]
	\centering
	\begin{adjustbox}{minipage=1.05\linewidth,scale=0.95}
		\begin{tabular}{|c|c|c|c|}
			\hline
			& 2d & 10d & 25d \\\hline
			CTSBN & $6.45 \pm 0.11$ & $3.83 \pm 0.20$ & $3.78 \pm 0.01$\\
			FCTSBN & $5.46 \pm 0.06$ & $3.43 \pm 0.03$ & $3.37 \pm 0.02$\\
			dFCTSBN & $\bm{5.09} \pm 0.11$ & $\bm{3.37} \pm 0.01$ & $\bm{3.35} \pm 0.02$ \\
			\hline
			FCRBM & $5.62 \pm 0.35$ & $3.77 \pm 0.38$ & $3.75 \pm 0.08$ \\\hline 
		\end{tabular}
		\caption{Average prediction error of 25 locations on the weather dataset.}
		\label{tab:weather}
	\end{adjustbox}
	\vspace{-4mm}
\end{table}
%

Average prediction error of 5 trials with random initializations are provided in Table \ref{tab:weather}. We observe that the factored model has significantly higher performance than the non-factored CTSBN, since the factored model has shared input-to-factor and factor-to-output parameters, and the factor-to-factor parameters have the potential to capture the correlations among side information at a higher level. Moreover, the performance improves as we increase the dimension of side information, which might be due to the increase of the number of parameters, along with the one-hot embeddings of side information. The 10d model has a significant performance boost compared with the 2d model, while using the 25d model further improves the performance slightly. We also compare our method with the one layer FCRBM model, which has 100 hidden variables and 50 variables for each of the three style factors and features, and achieved better results on the one-layer setting.

\subsection{Conditional Text Generation} \label{sec:text_gene}
Finally, we use a simple experiment to demonstrate the versatility of our model over the task of conditional text generation. We select 4 books (\emph{i.e.}, \textit{Napoleon the Little}, \textit{The Common Law}, \textit{Mysticism and Logic} and \textit{The Bible}) in the Gutenberg corpus, and use \emph{word2vec} \cite{mikolov2013distributed} to extract word embeddings, where words with few occurrences are omitted. This preprocessing step provides us with real-valued vectors, which are treated as observations and fit into our current model.


The real-valued word vector sequences are trained on a one-layer CTSBN of order 15 and 100 hidden variables. During generation, we clamp each generated value to the nearest possible word vector using cosine similarity. We focus on modeling real-valued sequences, hence the clamping approach is employed, instead of adopting a softmax layer to predict the words directly.

Table \ref{tab:text} displays the generated samples from the model when provided with the side information of \textit{The Bible} (B), \textit{The Common Law} (C) and an even mixture of these two books (BC).  As can be seen, our model learns to capture the styles associated with different books. 
By conditioning on the average of two styles, the model can generate reasonable samples that represent a hybrid of both styles (distinctive words such as \textit{\textless\#\textgreater} and \textit{god} from B, and \textit{application} and \textit{violence} from C), even though such
style combinations were not observed during training.

\begin{table}
	\centering
	\begin{adjustbox}{minipage=1.05\linewidth,scale=0.97}
		\small
		\begin{tabular}{|p{0.4cm}|p{7.2cm}|}
			\hline
			B & \textit{\textless\#\textgreater \ and 
				the evening and the morning were the fourth day . 
				\textless\#\textgreater \ and god said , let the waters 
				the heaven after his kind : \textless\#\textgreater \ god blessed 
				their thing months...}
			\\\hline
			
			
			C & \textit{we shall hold modern into tends ; and circumstance between seem 	
				understood retained defendant's to has that belief are not the recalled and will be led constituent...}\\\hline
			
			
			BC & 
			
			\textit{\textless\#\textgreater \ and unto the rendered fair violence 
				, morning turn the human whole been so eyes . 
				\textless\#\textgreater \ that god of air of the mountain show \textless\#\textgreater 
				the waters of fish and him would expect application : are gradual obliged that...}
			\\ \hline
			
		\end{tabular}
		\caption{Generated text. \textless\#\textgreater \ denote markers for verses.}
		\label{tab:text}
	\end{adjustbox}
\end{table}




%


\section{Conclusions and Future Work}
We have presented the Factored Conditional Temporal Sigmoid Belief Network, which can simultaneously model temporal sequences of multiple subjects.  A general framework for semi-supervised sequence classification is also provided, allowing one to train a conditional generative model along with a classifier. Experimental results on several datasets show that the proposed approaches obtain superior predictive performance, boost classification accuracy, and synthesize a large family of sequences.

While we assume side information as a simple vector, it would be interesting to incorporate more structured side information into the model, \emph{e.g.}, utilizing separate style and content components. We are also interested in models that can extract structured properties from side information.




\section*{Acknowledgements}
This research was supported in part by ARO, DARPA, DOE, NGA and ONR.
\bibliographystyle{icml2016}
\bibliography{fctsbn_icml}

\begin{thebibliography}{35}
\providecommand{\natexlab}[1]{#1}
\providecommand{\url}[1]{\texttt{#1}}
\expandafter\ifx\csname urlstyle\endcsname\relax
  \providecommand{\doi}[1]{doi: #1}\else
  \providecommand{\doi}{doi: \begingroup \urlstyle{rm}\Url}\fi

\bibitem[Bayer \& Osendorfer(2014)Bayer and Osendorfer]{bayer2014learning}
Bayer, J. and Osendorfer, C.
\newblock Learning stochastic recurrent networks.
\newblock In \emph{arXiv:1411.7610}, 2014.

\bibitem[Boulanger-Lewandowski et~al.(2012)Boulanger-Lewandowski, Bengio, and
  Vincent]{boulanger2012modeling}
Boulanger-Lewandowski, N., Bengio, Y., and Vincent, P.
\newblock Modeling temporal dependencies in high-dimensional sequences:
  Application to polyphonic music generation and transcription.
\newblock In \emph{ICML}, 2012.

\bibitem[Chung et~al.(2015{\natexlab{a}})Chung, Gulcehre, Cho, and
  Bengio]{chung2015gated}
Chung, J., Gulcehre, C., Cho, K., and Bengio, Y.
\newblock Gated feedback recurrent neural networks.
\newblock In \emph{ICML}, 2015{\natexlab{a}}.

\bibitem[Chung et~al.(2015{\natexlab{b}})Chung, Kastner, Dinh, Goel, Courville,
  and Bengio]{chung2015recurrent}
Chung, J., Kastner, K., Dinh, L., Goel, K., Courville, A., and Bengio, Y.
\newblock A recurrent latent variable model for sequential data.
\newblock In \emph{NIPS}, 2015{\natexlab{b}}.

\bibitem[Fabius et~al.(2014)Fabius, van Amersfoort, and
  Kingma]{fabius2014variational}
Fabius, O., van Amersfoort, J.~R., and Kingma, D.~P.
\newblock Variational recurrent auto-encoders.
\newblock In \emph{arXiv:1412.6581}, 2014.

\bibitem[Gammerman et~al.(1998)Gammerman, Vovk, and
  Vapnik]{gammerman1998learning}
Gammerman, A., Vovk, V., and Vapnik, V.
\newblock Learning by transduction.
\newblock In \emph{UAI}, 1998.

\bibitem[Gan et~al.(2015{\natexlab{a}})Gan, Chen, Henao, Carlson, and
  Carin]{gan2015scalable}
Gan, Z., Chen, C., Henao, R., Carlson, D., and Carin, L.
\newblock Scalable deep poisson factor analysis for topic modeling.
\newblock In \emph{ICML}, 2015{\natexlab{a}}.

\bibitem[Gan et~al.(2015{\natexlab{b}})Gan, Henao, Carlson, and
  Carin]{gan2015learning}
Gan, Z., Henao, R., Carlson, D., and Carin, L.
\newblock Learning deep sigmoid belief networks with data augmentation.
\newblock In \emph{AISTATS}, 2015{\natexlab{b}}.

\bibitem[Gan et~al.(2015{\natexlab{c}})Gan, Li, Henao, Carlson, and
  Carin]{TSBN_NIPS2015}
Gan, Z., Li, C., Henao, R., Carlson, D., and Carin, L.
\newblock Deep temporal sigmoid belief networks for sequence modeling.
\newblock In \emph{NIPS}, 2015{\natexlab{c}}.

\bibitem[Graves(2013)]{graves2013generating}
Graves, A.
\newblock Generating sequences with recurrent neural networks.
\newblock In \emph{arXiv:1308.0850}, 2013.

\bibitem[Hinton(2002)]{hinton2002training}
Hinton, G.~E.
\newblock Training products of experts by minimizing contrastive divergence.
\newblock In \emph{Neural computation}, 2002.

\bibitem[Hinton et~al.(1995)Hinton, Dayan, Frey, and Neal]{hinton1995wake}
Hinton, G.~E., Dayan, P., Frey, B.~J., and Neal, R.~M.
\newblock The ``wake-sleep'' algorithm for unsupervised neural networks.
\newblock In \emph{Science}, 1995.

\bibitem[Hinton et~al.(2006)Hinton, Osindero, and Teh]{hinton2006fast}
Hinton, G.~E., Osindero, S., and Teh, Y.~W.
\newblock A fast learning algorithm for deep belief nets.
\newblock In \emph{Neural Computation}, 2006.

\bibitem[Jordan et~al.(1999)Jordan, Ghahramani, Jaakkola, and
  Saul]{jordan1999an}
Jordan, M.~I., Ghahramani, Z., Jaakkola, T.~S., and Saul, L.~K.
\newblock An introduction to variational methods for graphical models.
\newblock In \emph{Machine Learning}, 1999.

\bibitem[Kingma \& Welling(2014)Kingma and Welling]{kingma2013auto}
Kingma, D.~P. and Welling, M.
\newblock Auto-encoding variational bayes.
\newblock In \emph{ICLR}, 2014.

\bibitem[Kingma et~al.(2014)Kingma, Mohamed, Rezende, and
  Welling]{kingma2014semi}
Kingma, D.~P., Mohamed, S., Rezende, D.~J., and Welling, M.
\newblock Semi-supervised learning with deep generative models.
\newblock In \emph{NIPS}, 2014.

\bibitem[Kiros et~al.(2014{\natexlab{a}})Kiros, Salakhutdinov, and
  Zemel]{kiros2014multimodal}
Kiros, R., Salakhutdinov, R., and Zemel, R.
\newblock Multimodal neural language models.
\newblock In \emph{ICML}, 2014{\natexlab{a}}.

\bibitem[Kiros et~al.(2014{\natexlab{b}})Kiros, Zemel, and
  Salakhutdinov]{kiros2014multiplicative}
Kiros, R., Zemel, R., and Salakhutdinov, R.~R.
\newblock A multiplicative model for learning distributed text-based attribute
  representations.
\newblock In \emph{NIPS}, 2014{\natexlab{b}}.

\bibitem[Le \& Mikolov(2014)Le and Mikolov]{le2014distributed}
Le, Q.~V. and Mikolov, T.
\newblock Distributed representations of sentences and documents.
\newblock In \emph{ICML}, 2014.

\bibitem[Li et~al.(2015)Li, Zhu, Shi, and Zhang]{li2015max}
Li, C., Zhu, J., Shi, T., and Zhang, Bo.
\newblock Max-margin deep generative models.
\newblock In \emph{NIPS}, 2015.

\bibitem[Liu et~al.(2010)Liu, Niculescu-Mizil, Lozano, and Lu]{liu2010learning}
Liu, Y., Niculescu-Mizil, A., Lozano, A.~C., and Lu, Y.
\newblock Learning temporal causal graphs for relational time-series analysis.
\newblock In \emph{ICML}, 2010.

\bibitem[Memisevic \& Hinton(2007)Memisevic and
  Hinton]{memisevic2007unsupervised}
Memisevic, R. and Hinton, G.
\newblock Unsupervised learning of image transformations.
\newblock In \emph{CVPR}, 2007.

\bibitem[Mikolov et~al.(2013)Mikolov, Sutskever, Chen, Corrado, and
  Dean]{mikolov2013distributed}
Mikolov, T., Sutskever, I., Chen, K., Corrado, G., and Dean, J.
\newblock Distributed representations of words and phrases and their
  compositionality.
\newblock In \emph{NIPS}, 2013.

\bibitem[Mittelman et~al.(2014)Mittelman, Kuipers, Savarese, and
  Lee]{mittelman2014structured}
Mittelman, R., Kuipers, B., Savarese, S., and Lee, H.
\newblock Structured recurrent temporal restricted boltzmann machines.
\newblock In \emph{ICML}, 2014.

\bibitem[Mnih \& Gregor(2014)Mnih and Gregor]{mnih2014neural}
Mnih, A. and Gregor, K.
\newblock Neural variational inference and learning in belief networks.
\newblock In \emph{ICML}, 2014.

\bibitem[Neal(1992)]{neal1992connectionist}
Neal, R.~M.
\newblock Connectionist learning of belief networks.
\newblock In \emph{Artificial intelligence}, 1992.

\bibitem[Rezende et~al.(2014)Rezende, Mohamed, and
  Wierstra]{rezende2014stochastic}
Rezende, D.~J., Mohamed, S., and Wierstra, D.
\newblock Stochastic backpropagation and approximate inference in deep
  generative models.
\newblock In \emph{ICML}, 2014.

\bibitem[Salakhutdinov \& Hinton(2009)Salakhutdinov and
  Hinton]{salakhutdinov2009deep}
Salakhutdinov, R. and Hinton, G.
\newblock Deep boltzmann machines.
\newblock In \emph{AISTATS}, 2009.

\bibitem[Srivastava et~al.(2015)Srivastava, Mansimov, and
  Salakhutdinov]{srivastava2015unsupervised}
Srivastava, N., Mansimov, E., and Salakhutdinov, R.
\newblock Unsupervised learning of video representations using lstms.
\newblock In \emph{ICML}, 2015.

\bibitem[Sutskever \& Hinton(2007)Sutskever and Hinton]{sutskever2007learning}
Sutskever, I. and Hinton, G.~E.
\newblock Learning multilevel distributed representations for high-dimensional
  sequences.
\newblock In \emph{AISTATS}, 2007.

\bibitem[Sutskever et~al.(2009)Sutskever, Hinton, and
  Taylor]{sutskever2009recurrent}
Sutskever, I., Hinton, G., and Taylor, G.
\newblock The recurrent temporal restricted boltzmann machine.
\newblock In \emph{NIPS}, 2009.

\bibitem[Sutskever et~al.(2011)Sutskever, Martens, and
  Hinton]{sutskever2011generating}
Sutskever, I., Martens, J., and Hinton, G.~E.
\newblock Generating text with recurrent neural networks.
\newblock In \emph{ICML}, 2011.

\bibitem[Taylor et~al.(2006)Taylor, Hinton, and Roweis]{taylor2006modeling}
Taylor, G., Hinton, G., and Roweis, S.
\newblock Modeling human motion using binary latent variables.
\newblock In \emph{NIPS}, 2006.

\bibitem[Taylor \& Hinton(2009)Taylor and Hinton]{taylor2009factored}
Taylor, G.~W. and Hinton, G.~E.
\newblock Factored conditional restricted boltzmann machines for modeling
  motion style.
\newblock In \emph{ICML}, 2009.

\bibitem[Tieleman \& Hinton(2012)Tieleman and Hinton]{tieleman2012lecture}
Tieleman, T. and Hinton, G.~E.
\newblock Lecture 6.5-rmsprop: Divide the gradient by a running average of its
  recent magnitude.
\newblock In \emph{COURSERA: Neural Networks for Machine Learning}, 2012.

\end{thebibliography}

\cleardoublepage
\appendix
\section{Variants of Conditional TSBNs}
In the main paper, we considered modeling real-valued sequence data. Other forms of data, such as binary and count data, can also be modeled by slight modification of the model.
\paragraph{Modeling binary data} Our models can be readily extended to model binary sequence data, by substituting $p(\vv_t | \vv_{t-1}, \hv_{t}, \yv_{t}) = \textrm{Ber}(\vv_t ; \sigma(\tilde{\vv}_t))$, where
\begin{align}
\tilde{\vv}_t = \Wmat_2^{(y)} \hv_t + \Wmat_4^{(y)} \vv_{t-1} + \cv^{(y)} 
\end{align}
$\sigma(x) = 1 / (1 + \exp (-x))$, and $\textrm{Ber}(x; p)$ denotes the Bernoulli distribution with parameter $p$.

\paragraph{Modeling count data} We also introduce an approach for modeling time-series data with count observations, $p(\vv_t | \vv_{t-1}, \hv_t, \yv_t) = \prod_{m=1}^{M} s_{mt}^{v_{mt}}$, where
\begin{align}
s_{mt} = \frac{\exp(\hv_t^\top \wv_{2m}^{(y)}  + \vv_{t-1}^\top \wv_{4m}^{(y)} + c_m^{(y)})  }{\sum_{m^\prime = 1}^{M} \exp(\hv_t^\top \wv_{2m^\prime}^{(y)}  + \vv_{t-1}^\top \wv_{4m^\prime}^{(y)} + c_{m^\prime}^{(y)})}
\end{align}

%
\begin{algorithm}
	\centering
	\caption{Calculate gradient estimates for model parameters and recognition parameters.}
	\label{alg:nvil_gradients}
	\begin{algorithmic}
		\STATE $\Delta \thetav \gets \bm{0}$, $\Delta \phiv \gets \bm{0}$, $\Delta \lambdav \gets \bm{0}$
		\STATE $\kappa \gets 0$, $\tau \gets 0$
		\STATE $\Lcal \gets \bm{0}$
		\FOR{$t \gets 1$ \textbf{to} $T$} 
		\STATE $\hv_t \sim q_\phi (\hv_t | \vv_t)$
		\STATE $l_t \gets \log p_\theta (\vv_t, \hv_t) - \log q_\theta (\hv_t | \vv_t)$
		\STATE $\Lcal \gets \Lcal + l_t$
		\STATE $l_t \gets l_t - C_\lambdav (\vv_t)$
		\ENDFOR 
		\STATE $\kappa_b \gets \mathrm{mean}(l_1, \ldots, l_T)$
		\STATE $\tau_b \gets \mathrm{var}(l_1, \ldots, l_T)$
		\STATE $\kappa \gets \rho \kappa + (1 - \rho) \kappa_b$
		\STATE $\tau \gets \rho \tau + (1 - \rho) \tau_b$
		\FOR{$t \gets 1$ \textbf{to} $T$} 
		\STATE $l_t \gets \frac{l_t - \kappa}{\max (1, \sqrt{\tau})}$
		\STATE $\Delta \thetav \gets \Delta \thetav + \nabla_{\thetav} \log p_{\thetav} (\vv_t, \hv_t)$
		\STATE $\Delta \phiv \gets \Delta \phiv + l_t \nabla_{\phiv} \log q_{\phiv} (\hv_t | \vv_t)$
		\STATE $\Delta \lambdav \gets \Delta \lambdav + l_t \nabla_\lambdav C_\lambdav (\vv_t)$
		\ENDFOR
	\end{algorithmic}
\end{algorithm}

\vspace{-1.5mm}
\section{Inference Details}
\vspace{-0.5mm}
\subsection{Outline of NVIL Algorithm}
The outline of the NVIL Algorithm for computing gradients are shown in Algorithm \ref{alg:nvil_gradients}. $C_\lambdav (\vv_t)$ represents the sum of data-dependent baseline and data-independent baseline.

\vspace{-1.0mm}
\subsection{Derivatives for Conditional TSBNs}
\vspace{-1.0mm}
For the Conditional TSBNs, we have:
\begin{gather}
p(h_{jt}=1|\hv_{t-1}, \vv_{t-1}, \yv_{t})= \sigma(\tilde{h}_{jt})   \\
p(\vv_t|\hv_t, \vv_{t-1}, \yv_{t}) = \Ncal(\muv_t, \mbox{diag}(\sigmav_t^2)) \\
\tilde{\hv}_t = \Wmat_{1}^{(y)} \hv_{t-1} + \Wmat_{3}^{(y)} \vv_{t-1} + \bv^{(y)}  \\
\muv_t = \Wmat_{2}^{(y)} \hv_t  + \Wmat_{4}^{(y)} \vv_{t-1} + \cv^{(y)}   \\
\log\sigmav_t^2 = \Wmat_{2}^{\prime(y)} \hv_t  + \Wmat_{4}^{\prime(y)} \vv_{t-1} + \cv^{\prime(y)} 
\end{gather}
The recognition model is expressed as:
\begin{align}
q(h_{jt} = 1|\hv_{t-1}, \vv_t, \vv_{t-1}, \yv_t) = \sigma(\hat{h}_{jt})  \\
\hat{\hv}_t = \Umat_1^{(y)} \hv_{t-1} + \Umat_2^{(y)} \vv_{t} + \Umat_3^{(y)} \vv_{t-1} + \dv^{(y)} 
\end{align}
In order to implement the NVIL algorithm, we need to calculate the lower bound and also the gradients. Specifically, the lower bound can be expressed as $\Lcal = \sum_{t=1}^{T} \E_{q_\phi (\Hmat | \Vmat)} [l_t]$, where
\begin{align}
l_t &= \sum_{j=1}^{J}( \tilde{h}_{jt} h_{jt} - \log (1 + \exp (\tilde{h}_{jt})) ) \nonumber \\
&+ \sum_{m=1}^{M} ( \frac{1}{2} \log 2\pi + \log \sigma_{mt} + \frac{(v_{mt} - \mu_{mt})^2}{2 \sigma_{mt}^2})  \nonumber \\
&+ \sum_{j=1}^{M} ( \hat{h}_{jt} h_{jt} - \log (1 + \exp (\hat{h}_{jt})))
\end{align}
The gradients for model parameters $\theta$ are 
\begin{align}
\frac{\partial \log p_\theta(\vv_t, \hv_t)}{\partial \hat{\Wmat}_{1jj^\prime s}} &= (h_{jt} - \sigma(\tilde{h}_{jt})) \cdot h_{j^\prime (t-1)} \cdot y_{st}  \label{eq:w1_grad} \\
\frac{\partial \log p_\theta(\vv_t, \hv_t)}{\partial \hat{\Wmat}_{2mjs}} &= \frac{v_{mt} - \mu_{mt}}{\sigma_{mt}^2} \cdot h_{jt} \cdot y_{st} \\
\frac{\partial \log p_\theta(\vv_t, \hv_t)}{\partial \hat{\Wmat}_{2mjs}^\prime} &= (\frac{(v_{mt} - \mu_{mt})^2}{\sigma_{mt}^2} - 1) \cdot h_{jt} \cdot y_{st}  \\
\frac{\partial \log p_\theta(\vv_t, \hv_t)}{\partial \hat{\Wmat}_{3jms}} &= (h_{jt} - \sigma(\tilde{h}_{jt})) \cdot v_{m (t-1)} \cdot y_{st}  \\
\frac{\partial \log p_\theta(\vv_t, \hv_t)}{\partial \hat{\Wmat}_{4mm^\prime s}} &= \frac{v_{mt} - \mu_{mt}}{\sigma_{mt}^2} \cdot v_{m^\prime (t-1)} \cdot y_{st}  \\
\frac{\partial \log p_\theta(\vv_t, \hv_t)}{\partial \hat{\Wmat}_{4mm^\prime s}^\prime} &= (\frac{(v_{mt} - \mu_{mt})^2}{\sigma_{mt}^2} - 1) \cdot v_{m^\prime (t-1)} \cdot y_{st}  \\
\frac{\partial \log p_\theta(\vv_t, \hv_t)}{\partial \Bmat_{js}} &= (h_{jt} - \sigma(\tilde{h}_{jt})) \cdot y_{st}  \label{eq:b_grad}\\
\frac{\partial \log p_\theta(\vv_t, \hv_t)}{\partial \Cmat_{ms}} &= (\frac{v_{mt} - \mu_{mt}}{\sigma_{mt}^2} \cdot y_{st}) \cdot y_{st} \label{eq:c_grad}  \\
\frac{\partial \log p_\theta(\vv_t, \hv_t)}{\partial \Cmat_{ms}^\prime} &= (\frac{(v_{mt} - \mu_{mt})^2}{\sigma_{mt}^2} - 1) \cdot y_{st} \label{eq:c_prime_grad} 
\end{align}
The gradients for recognition parameters $\phi$ are
\begin{align}
\frac{\partial \log q_\phi (\hv_t | \vv_t)}{\partial \hat{\Umat}_{1jj^\prime s}} &= (h_{jt} - \sigma(\hat{h}_{jt})) \cdot h_{j^\prime (t-1)} \cdot y_{st}  \label{eq:u1_grad}\\
\frac{\partial \log q_\phi (\hv_t | \vv_t)}{\partial \hat{\Umat}_{2jm s}} &= (h_{jt} - \sigma(\hat{h}_{jt})) \cdot h_{mt} \cdot y_{st}  \\
\frac{\partial \log q_\phi (\hv_t | \vv_t)}{\partial \hat{\Umat}_{3jm s}} &= (h_{jt} - \sigma(\hat{h}_{jt})) \cdot v_{m (t-1)} \cdot y_{st}  \\
\frac{\partial \log q_\phi (\hv_t | \vv_t)}{\partial \hat{\Dmat}_{js}} &= (h_{jt} - \sigma(\hat{h}_{jt})) \cdot y_{st} \label{eq:d_grad} 
\end{align}

\subsection{Derivatives for the Factored Model}
The factored model substitutes the weight tensors $\hat{\Wmat}$ with three matrices $\Wmat_a$, $\Wmat_b$, and $\Wmat_c$, such that
\begin{align}
\Wmat^{(y)} = \Wmat_a \cdot \textrm{diag}(\Wmat_b \yv_t) \cdot \Wmat_c 
\end{align}
We notice that for a particular $\Wmat^{(y)}$, the objective function for the $t$-th time step can be generalized as $\Lcal^\prime = f(\Wmat^{(y)} \etav + \chiv ) + \rho $, where the elements of $\frac{\partial \etav}{\Wmat^{(y)}}$,  $\frac{\partial \chiv}{\Wmat^{(y)}}$ and $\frac{\partial \rho}{\Wmat_{(y)}}$ are zero. Assuming $\xiv = f^\prime(\Wmat^{(y)} \etav + \chiv)$, we have the following gradients:
\begin{align}
\frac{\partial \Lcal^\prime}{\partial \Wmat_a} &= \xiv \cdot (\textrm{diag}(\Wmat_b \yv_t) \cdot \Wmat_c \etav)^\top \label{eq:wa_grad}  \\
\frac{\partial \Lcal^\prime}{\partial \Wmat_b} &= ((\Wmat_a^\top \xiv) \odot (\Wmat_c \etav)) \cdot \yv_t^\top \label{eq:wb_grad}  \\
\frac{\partial \Lcal^\prime}{\partial \Wmat_c} &= \textrm{diag}(\Wmat_b \yv_t) \cdot \Wmat_a \cdot \xiv \etav^\top \label{eq:wc_grad} 
\end{align}
where $\Amat \odot \Bmat$ denotes the element-wise product between matrices $\Amat$ and $\Bmat$ with the same dimensions.

For FCTSBN, the gradients for bias parameters in \ref{eq:b_grad}, \ref{eq:c_grad}, \ref{eq:c_prime_grad} and \ref{eq:d_grad} remains the same, while gradients for the factored weight parameters can be calculated using \ref{eq:wa_grad} - \ref{eq:wc_grad}. For $\Wmat_{1a}$, $\Wmat_{1b}$ and $\Wmat_{1c}$, we have $\xiv = \hv_{t} - \sigma(\tilde{\hv}_{t})$ and $\etav = \hv_{t-1}$. Hence the gradients for these parameters are:
\begin{adjustbox}{minipage=1.05\linewidth,scale=0.91}
	\begin{align}
	\frac{\partial \log p_\theta(\vv_t, \hv_t)}{\partial \Wmat_{1a}} &= (\hv_{t} - \sigma(\tilde{\hv}_{t})) \cdot (\mathrm{diag}(\Wmat_{1b} \yv_t) \cdot \Wmat_{1c} \hv_{t-1})^\top \\
	\frac{\partial \log p_\theta(\vv_t, \hv_t)}{\partial \Wmat_{1b}} &= ((\Wmat_{1a}^\top (\hv_{t} - \sigma(\tilde{\hv}_{t}))) \odot (\Wmat_{1c} \hv_{t-1})) \cdot \yv_t^\top \\
	\frac{\partial \log p_\theta(\vv_t, \hv_t)}{\partial \Wmat_{1c}} &= \mathrm{diag}(\Wmat_{1b} \yv_t) \cdot \Wmat_{1a} \cdot (\hv_{t} - \sigma(\tilde{\hv}_{t})) \hv_{t-1}^\top
	\end{align} 
\end{adjustbox}
Gradients of other factored parameters can be derived in analogy.

\subsection{Gradients for Deep Models}
Suppose we have a two-layered deep CTSBN with the joint probability distribution $p_\theta(\Vmat, \Hmat, \Zmat)$. $\Zmat = [\zv_1, \cdot , \zv_T]$, where $\zv_t \in \{0, 1\} ^ K$ is the latent variables on the second layer at time $t$. The probability of $\zv_t$ and $\hv_t$ are characterized by
\begin{gather}
p(z_{kt} = 1| \hv_{t-1}, \zv_{t-1}, \yv_t) = \sigma(\tilde{z}_{et}) \\
p(h_{jt} = 1| \hv_{t-1}, \zv_{t}, \vv_{t-1}, \yv_t) = \sigma(\tilde{h}_{jt}) \\
\tilde{\zv}_v = \Wmat_6^{(y)} \hv_{t-1} + \Wmat_7^{(y)} \zv_{t-1} + \av^{(y)} \\
\tilde{\hv}_t = \Wmat_1^{(y)} \hv_{t-1} + \Wmat_3^{(y)} \vv_{t-1} + \Wmat_5^{(y)} \zv_{t} + \bv^{(y)}
\end{gather}
The recognition model for $\zv_t$ is expressed as:
\begin{align}
q(z_{kt} = 1|\zv_{t-1}, \hv_t, \hv_{t-1}, \yv_t) = \sigma(\hat{z}_{kt}) \\
\hat{\zv}_t = \Umat_4^{(y)} \zv_{t-1}  + \Umat_{5}^{(y)} \hv_{t} + \Umat_6^{(y)} \hv_{t-1} + \ev^{(y)}
\end{align}

The lower bound takes the form $\Lcal = \sum_{t=1}^{T} \E_{q_\phi(\Zmat, \Hmat | \Vmat)} [l_t]$, where
\begin{align}
l_t &= \sum_{j=1}^{J}( \tilde{h}_{jt} h_{jt} - \log (1 + \exp (\tilde{h}_{jt})) ) \nonumber \\
&+ \sum_{m=1}^{M} ( \frac{1}{2} \log 2\pi + \log \sigma_{mt} + \frac{(v_{mt} - \mu_{mt})^2}{2 \sigma_{mt}^2}) \nonumber \\
&+ \sum_{j=1}^{M} ( \hat{h}_{jt} h_{jt} - \log (1 + \exp (\hat{h}_{jt}))) \nonumber \\
&+ \sum_{k=1}^{K}( \tilde{z}_{kt} z_{kt} - \log (1 + \exp (\tilde{z}_{kt})) ) \nonumber \\
&+ \sum_{k=1}^{K}( \hat{z}_{kt} z_{kt} - \log (1 + \exp (\hat{z}_{kt})) )
\end{align}
Compared with the one-layer model, the two-layer model adds two terms concerning $\zv_t$, whose form is similar to that of $\hv_t$. Therefore, gradients for additional parameters ($\hat{\Wmat}_5$, $\hat{\Wmat}_6$, $\hat{\Wmat}_7$, $\hat{\Umat}_4$, $\hat{\Umat}_5$, $\hat{\Umat}_6$, $\Amat$ and $\Emat$) can be readily calculated using Equations \ref{eq:w1_grad}, \ref{eq:u1_grad}, \ref{eq:b_grad}. If the model parameters are factored, gradients can be calculated using \ref{eq:wa_grad}-\ref{eq:wc_grad}.
\subsection{Semi-supervised FCTSBN}
The lower bound for semi-supervised learning $\Lcal_s$ can be described as $\Lcal_s = \Lcal_l + \Lcal_u$, where $\Lcal_l$ and $\Lcal_u$ are the lower bounds for labeled data and unlabeled data respectively:
\begin{align}
\Lcal_l = \Lcal + \alpha \cdot \E_{\tilde{p}_l(\Vmat, \Ymat)}[\log q_{\thetav}(\Ymat|\Vmat)] \\
\Lcal_u = \mathcal{J}(q_{\phiv}(\Hmat, \Ymat|\Vmat), p_{\thetav} (\Hmat, \Vmat, \Ymat)) 
\end{align}

For labeled data, $\Lcal_l$ adds a classification loss to the generative model lower bound, where the gradient can be readily calculated from Algorithm \ref{alg:nvil_gradients} plus a term $\alpha \nabla_{\thetav} \E_{\tilde{p}_l(\Vmat, \Ymat)}[\log q_{\thetav}(\Ymat|\Vmat)]$, which can also be approximated using Monte-Carlo integration. For unlabeled data, $\Lcal_u$ requires calculating the expectation with respect to $q_{\phiv}(\Hmat, \Ymat | \Vmat) = q_{\phiv}(\Hmat | \Vmat, \Ymat) q_{\phiv}(\Ymat | \Vmat)$. Hence, to optimize $\Lcal_u$, we can sample $\Ymat$ from $\Vmat$ first, and then apply Algorithm \ref{alg:nvil_gradients} to obtain the approximated gradients. Outline for optimizing $\Lcal_s$ is shown in Algorithm \ref{alg:semi_learning}.

\begin{algorithm}
	\centering
	\caption{Optimizing the semi-supervised objective $\Lcal_s$.}
	\label{alg:semi_learning}
	\begin{algorithmic}
		\STATE Initialize $\thetav$, $\phiv$
		\WHILE{not converged}
		\IF{sample labeled data}
		\STATE $(\vv, \yv) \sim \tilde{p}_l(\Vmat, \Ymat)$
		\ELSE
		\STATE $\vv \sim \tilde{p}_u(\Vmat)$
		\STATE $\yv \sim q_{\phiv}(\yv | \vv)$
		\ENDIF
		\STATE Calculate $\nabla_{\thetav} \Lcal_s$ and $\nabla_{\phiv} \Lcal_s$
		\STATE $\thetav \gets \thetav + \nabla_{\thetav} \Lcal_s$
		\STATE $\phiv \gets \phiv + \nabla_{\phiv} \Lcal_s$
		\ENDWHILE
	\end{algorithmic}
\end{algorithm}

\subsection{Computational Complexity}
Although side information is included, the computational complexity of learning and inference for CTSBN and FCTSBN are comparable to that of TSBN. Suppose that $\yv_t$ are one-hot encoded vectors, so gradients of some parameters are zero and does not cost computation. For CTSBN, the complexity for calculating the gradient would be $O\big((J + M)^2 m n \big)$, where $m$ is the size of the mini batch, and $n$ is the order. For FCTSBN, the complexity would be $O\big(F (J + M) m n \big)$.

Empirically, training for even the largest model in our experiments takes around 10 hours on unoptimized MATLAB using a laptop, whereas generating thousands of samples can be achieved within seconds.
\begin{table}[t!]
	\centering
	\begin{tabular}{|c|c|c|c|}
		\hline
		Style & FCTSBN & dFCTSBN & FCRBM   \\\hline
		Sexy & 0.4971 & 0.2401 & 0.4926  \\
		Strong & 0.2899 & 0.2415 & 0.3385  \\
		Cat & 0.1858 & 0.1732 & 0.3475  \\
		Dinosaur & 0.6299 & 0.4182 & 0.3387  \\
		Drunk & 0.6227 & 0.6184 & 0.5005  \\
		Gangly & 0.5553 & 0.3777 & 0.5474  \\
		Chicken & 0.7798 & 0.6909 & 0.3519  \\
		Graceful & 0.7184 & 0.4232 & 0.3544  \\
		Normal & 0.3043 & 0.2330 & 0.2713  \\
		Old man & 0.2483 & 0.1831 & 0.8125  \\\hline\hline
		Average & 0.4832 & 0.3599 & 0.4355  \\\hline
	\end{tabular}
	\caption{Average prediction error for mocap10}
	\label{tab:mocap10_error}
\end{table}
\begin{table*}[t!]
	\centering
	\begin{tabular}{|c|c|c|c|c|c|c|}
		\hline
		\% of labeled data & 10-KNN & Softmax-0.001 & Softmax-1 & TSVM & FCTSBN & dFCTSBN \\\hline
		0.25 & 73.16 & $72.07 \pm 0.52$ & $71.44 \pm 0.55$ & 76.53 & $78.33 \pm 1.75$ & $79.95 \pm 2.04$ \\\hline
		0.3125 & 78.00 & $78.11 \pm 0.21$ & $78.07 \pm 1.34$ & 79.47 & $81.11 \pm 1.22$ & $82.94 \pm 1.08$ \\\hline
		0.375 & 81.37 & $82.31 \pm 0.58$ & $80.53 \pm 0.73$ & 81.47 & $83.74 \pm 1.03$ & $86.10 \pm 1.53$ \\\hline
		0.4375 & 83.68 & $83.26 \pm 0.84$ & $80.91 \pm 2.06$ & 82.22 & $84.37 \pm 2.96$ & $87.20 \pm 1.28$ \\\hline
		0.5 & 84.53 & $83.81 \pm 0.86$ & $81.26 \pm 1.42$ & 83.37 & $85.17 \pm 1.84$ & $87.70 \pm 2.16$ \\\hline
		0.5625 & 86.00 & $85.11 \pm 2.90$ & $81.28 \pm 1.64$ & 84.31 & $86.84 \pm 1.18$ & $88.16 \pm 1.25$ \\\hline
		0.625 & 85.47 & $85.78 \pm 0.95$ & $84.94 \pm 0.74$ & 86.94 & $88.17 \pm 1.66$ & $89.40 \pm 1.55$ \\\hline
		0.6875 & 86.00 & $86.03 \pm 1.49$ & $83.95 \pm 1.26$ & 86.63 & $88.63 \pm 1.45$ & $89.40 \pm 1.96$ \\\hline
		0.75 & 84.74 & $85.98 \pm 1.51$ & $85.07 \pm 0.88$ & 87.26 & $88.73 \pm 1.68$ & $89.57 \pm 1.89$ \\\hline
	\end{tabular}
	\caption{Test accuracy (in percentage) of mocap10 classification.}
	\label{tab:mocap10_classification}
\end{table*}
\section{Extended Experiment Results}
\subsection{Generated Videos}
Along with this supplementary article including more details for our model, we present a number of videos to demonstrate the generative capacities of our models. The videos are available at \href{https://goo.gl/9R59d7}{https://goo.gl/9R59d7}.

\paragraph{mocap2} We present synthesized sequences by the semi-supervised Hidden Markov FCTSBN trained with labeled and unlabeled data, namely \textbf{walk.mp4}, \textbf{run.mp4}, \textbf{walk-run.mp4} and \textbf{run-walk.mp4}. 

The videos denotes sequences of (\textit{i}) walking; (\textit{ii}) running; (\textit{iii}) transition from walking to running; and (\textit{iv}) transition from running to walking.

\paragraph{mocap10} Sequences produced by the Hidden Markov FCTSBN over the mocap10 dataset are presented, including 9 styles and some style transitions and combinations.

\subsection{Detailed Results on mocap10 Prediction}
Average prediction error for each style can be found in Table \ref{tab:mocap10_error}. For the FCTSBN and deep FCTSBN, each hidden layer has 100 hidden variables and 50 factors, whereas the FCRBM contains 600 hidden variables, 200 features and 200 factors.

\subsection{Detailed Results on mocap10 Classification}
We include error bar results for the mocap10 classification task in Table \ref{tab:mocap10_classification}.

\end{document}